\documentclass[a4paper,fleqn]{cas-dc}

\usepackage[numbers]{natbib}
\usepackage{lineno}
\usepackage{subfigure}
\usepackage{algorithm,algpseudocode}
\usepackage{times}
\usepackage{epsfig}
\usepackage{graphicx}
\usepackage{amsmath}
\usepackage{amssymb}
\usepackage{booktabs} 
\usepackage{amsthm}
\usepackage{diagbox}
\usepackage{caption}
\usepackage{float}
\usepackage{epstopdf}
\usepackage{enumitem}
\usepackage{multirow}
\usepackage{color}
\newtheorem{theorem}{Theorem}[section]

\def\tsc#1{\csdef{#1}{\textsc{\lowercase{#1}}\xspace}}
\tsc{WGM}
\tsc{QE}
\tsc{EP}
\tsc{PMS}
\tsc{BEC}
\tsc{DE}

\begin{document}
	\let\WriteBookmarks\relax
	\def\floatpagepagefraction{1}
	\def\textpagefraction{.001}
	\shorttitle{Leveraging social media news}
	\shortauthors{CV Radhakrishnan et~al.}
	
	\title [mode = title]{Hard Class Rectification for Domain Adaptation}

	\author[1]{Yunlong Zhang}[style=chinese]
	\author[1]{Changxing Jing}[style=chinese]
	\author[1]{Huangxing Lin }[style=chinese]
	\author[1]{Chaoqi Chen}[style=chinese]
	\author[1]{Yue Huang}[style=chinese]
	\ead{yhuang2010@xmu.edu.cn}
	\cormark[1]
	\author[1]{Xinghao Ding}[style=chinese]
	\author[2]{Yang Zou}[style=chinese]

	\address[1]{Fujian Key Laboratory of Sensing and Computing for SmartCity, School of Informatics, Xiamen University, Xiamen, Fujian, 361005, China}
	\address[2]{Electrical and Computer Engineering, Carnegie Mellon University, Pittsburgh, PA, 15213, U.S.A.}

	\begin{abstract}
		Domain adaptation (DA) aims to transfer knowledge from a label-rich and related domain (source domain) to a label-scare domain (target domain).  Pseudo-labeling has recently been widely explored and used in DA. However, this line of research is still confined to the inaccuracy of pseudo labels. In this paper, we explore the imbalance issue of performance among classes in-depth and observe that the worse performances among all classes are likely to further deteriorate in the pseudo-labeling, which not only harms the overall transfer performance but also restricts the application of DA. In this paper,  we propose a novel framework, called Hard Class Rectification Pseudo-labeling (HCRPL), to alleviate this problem from two aspects. First, we propose a simple yet effective scheme, named Adaptive Prediction Calibration (APC), to calibrate predictions of target samples. Then, we further consider the predictions of calibrated ones, especially for those belonging to the hard classes, which are vulnerable to perturbations. To prevent these samples to be misclassified easily, we introduce Temporal-Ensembling (TE) and Self-Ensembling (SE) to obtain consistent predictions. The proposed method is evaluated on both unsupervised domain adaptation (UDA) and semi-supervised domain adaptation (SSDA). Experimental results on several real-world cross-domain benchmarks, including ImageCLEF, Office-31, Office+Caltech, and Office-Home, substantiate the superiority of the proposed method.
	\end{abstract}


	\maketitle
	
	\section{Introduction}
	
	\label{sec1}
	
	Over the last few years, Deep Neural Networks (DNNs) \cite{lecun2015deep} achieved impressive performance in machine learning tasks, such as computer vision \cite{he2016deep}, speech recognition \cite{Amodei2015Deep}, medical analysis \cite{Zhang2020CollaborativeUD}, {industrial fault diagnosis \cite{Li2019WaveletKernelNetAI,Li2020MultireceptiveFG},} and so on. Nevertheless, collecting and annotating large-scale training data in distinct domains for various applications is an expensive and labor-intensive process. Meanwhile, the application of DNNs is greatly limited because the learned network shows poor generalization ability when it encounters new environments. Domain adaptation (DA) \cite{pan2009survey} serves as an ideal solution for addressing this problem. It has raised widespread attentions \cite{DBLP:journals/ml/Ben-DavidBCKPV10,ganin2016domain} in the machine learning community.

	The majority of existing DA methods \cite{gretton2012kernel,long2015learning,ganin2014unsupervised,ganin2016domain,SunS16Deep,SanjayVariational,tzeng2014deep,sun2016return,shao2018feature} were devoted to aligning source and target features by decreasing domain divergence, and these methods can be supported by the theoretical analysis of DA  \cite{DBLP:journals/ml/Ben-DavidBCKPV10}. However, there are still two main limitations with these approaches: 1) the methods of global alignment of the source and target features cannot guarantee correct alignment of class-level representations, and 2) the global alignment methods cannot learn target-discriminative representations. Aligning the class conditional distributions of the source and target domains is an effective tool to tackle these limitations. However, directly pursuing the alignment of class conditional distributions is impossible due to the absence of target labels.
	
	\begin{figure}[!t]
		\centering
		\includegraphics[width=0.85\columnwidth]{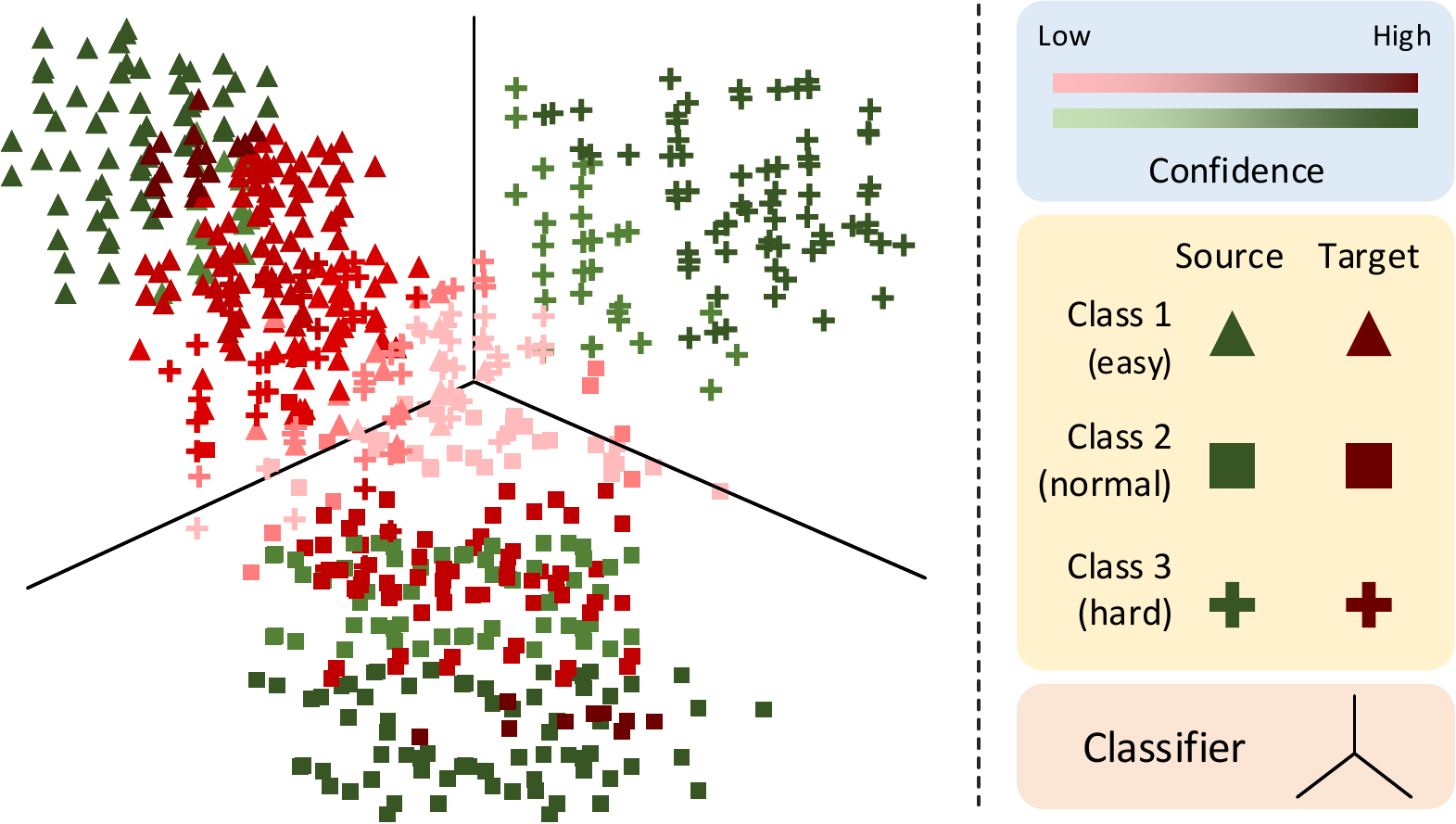}
		\caption{Hard class problem in existing pseudo-labeling based DA methods: Compared with class 1 and 2, class 3 has lower predictive class proportion (i.e., (the number of samples classified into a certain class)/(the number of target samples)). Meanwhile, for this class, target samples with higher confidence are mainly classified into class 1.
		}\label{fig1}
	\end{figure}
	
	Pseudo-labeling \cite{lee2013pseudo} was first employed for semi-super-vised learning tasks. Recently, it was also introduced into DA to solve the aforementioned limitations by alternatively selecting the target samples with high confident predictions as the pseudo-labeled target set (labeling phase) and training the model with the source domain and pseudo-labeled target set (training phase). Although pseudo-labeling is considered to be a promising paradigm, it is still limited due to inevitable false pseudo labels. Zhang et al. \cite{zhang2016understanding}  demonstrated that false labels are easily fit by DNN, which harms its generalization. Retraining DNN with the false pseudo-labeled samples does not guarantee the generalization ability of the target domain. We further analyze the theory of DA \cite{DBLP:journals/ml/Ben-DavidBCKPV10} in Section \ref{sec_hard_class} and demonstrate that the expected error on the target domain is determined by false pseudo labels ratio (i.e., (the number of incorrectly pseudo-labeled samples)/(the number of pseudo-labeled samples)). Therefore, reducing the false pseudo labels ratio is of crucial importance to pseudo-labeling methods. {To achieve this}, Zou et al. \cite{zou2018domain} enhanced pseudo-labeling from two aspects. 1) They introduced self-paced learning which generates pseudo labels from easy to hard to alleviate error accumulation of pseudo labels. 2) They utilized different confidence thresholds to select a target pseudo-labeled set for different classes. Zou et al. \cite{zou2019confidence} then introduced confidence regularization to avoid overconfident labels. Saito et al. \cite{saito2017asymmetric} adopted two classifiers with a multiview loss to label the target samples and used a fixed confidence threshold to pick up the reliable pseudo labels. Some works \cite{xie2018learning,chen2019progressive,DBLP:journals/corr/abs-1812-00893,chen2019joint} generate pseudo labels in the feature space and adopt different distances to measure confidence. 
	
	In this paper, we explore a problem that is unconsidering in the aforesaid methods. As shown in Figure \ref{fig1}, the classifier is trained on the source domain, and three classes deliver distinct performances. Class 1 belongs to easy classes and has a higher predictive class proportion (i.e., (the number of samples classified into a certain class)/(the number of target samples)). The target samples belonging to these classes are very likely to be classified correctly without extra manipulation. The classifier trained on the source domain can well generalize to the target domain for these classes. Class 2 belongs to normal classes and has a moderate predictive class proportion. Although some target samples belonging to it are misclassified, the predictions with higher confidence have higher accuracy. Therefore, the existing pseudo-labeling methods \cite{zou2018domain,zou2019confidence,saito2017asymmetric,xie2018learning,chen2019progressive,DBLP:journals/corr/abs-1812-00893,chen2019joint} progressively improve the performance on these classes by adding the samples with higher confident predictions into training. Class 3 belongs to hard classes and has a lower predictive class proportion. Meanwhile, for hard classes, even the target samples with higher confident predictions also are highly possible to be misclassified, adding these samples into training will misguide the classifier. Therefore, the pseudo-labeling methods cannot improve the performance of the hard classes and even deteriorate it, which is the main difference between the normal and hard classes.
	
	To tackle the hard class problem, we present a simple yet effective scheme, named Adaptive Prediction Calibration (APC), and it calibrates the predictions of target samples to promote the hard classes, to maintain the normal ones, and to attenuate the easy ones. Furthermore, we consider the calibrated predictions are unstable and unreliable in the pseudo-labeling from the following aspects. First, DNNs are vulnerable to target samples since they are far away from the source domain \cite{DBLP:journals/corr/SzegedyZSBEGF13} {(i.e., distributional mismatch)}. Despite encountering a small perturbation (e.g., different augmentations, different classifiers), the predictions of target samples are changed drastically. Second, the APC will further magnify and deteriorate the unrobustness for hard classes since their predictions are enlarged proportionally {(i.e., deviation magnification)}. To ensure the reliability of calibrated predictions,  we propose two ensembling methods, Temporal-Ensembling (TE) and Self-Ensembling (SE).
	
	The proposed schemes can be directly combined with the existing pseudo-labeling methods. In this paper, based on CBST  \cite{zou2018domain}, we propose a novel pseudo-labeling framework, which combines APC, SE, and TE to alleviate the hard class problem. The proposed framework is called Hard Class Rectification Pseudo-labeling (HCRPL).

	The main contributions of this work can be summarized as follows:
	\begin{itemize}
		\item [i)]
		{We reveal the hard class problem in DA, which harms the performance of pseudo-labeling and restricts its applications. }
		\item [ii)]
		{We propose a calibration method (i.e., APC) to alleviate the hard class problem by promoting the hard classes, maintaining the normal ones, and attenuating the easy ones. Furthermore, the predictions of target samples are vulnerable because of the distributional mismatch and deviation magnification. Hence,  we introduce TE and SE to improve the reliability of predictions.} 
		\item [iii)]
		We evaluate HCRPL on four public datasets under both UDA and SSDA settings. Extensive experimental results show that the proposed method achieves promising results in various DA scenarios. 
	\end{itemize}
	
	\section{Related Work} \label{sec3}
	\subsection{Unsupervised Domain Adaptation} 
	Under the UDA setting,  we are given a set of labeled source samples and a set of unlabeled target examples. According to Ben-David et al.'s \cite{DBLP:journals/ml/Ben-DavidBCKPV10} theoretical analysis of DA, the expected error for the target domain depends on three terms: expected error on the source domain, domain divergence, and shared error of ideal joint hypothesis, can divide UDA methods into two parts. In the first part, researchers assumed that the shared error of the ideal joint hypothesis was small and mainly focused on decreasing the domain divergence. One method of aligning distributions is through minimizing statistical divergences that measure the distance between two distributions. Representative divergences mainly include Maximum mean discrepancy (MMD) \cite{gretton2012kernel}, Correlation alignment (CORAL) \cite{sun2016return}, Contrastive domain discrepancy (CDD) \cite{kang2019contrastive}, and so on. Inspired by GAN \cite{goodfellow2014generative}, numerous methods \cite{ganin2014unsupervised,ganin2016domain,SunS16Deep,SanjayVariational,tzeng2014deep,sun2016return,shao2018feature,chen2020harmonizing,chen2021I3NET} were proposed  to align source and target domains by adversarial training. Although domain divergence was decreased, the shared error of the ideal joint hypothesis would be large if class conditional distributions are not aligned and separated. In the second part, researchers paid more attention to decreasing the shared error of the ideal joint hypothesis. 
	
	Among many schemes, pseudo-labeling is a promising paradigm for reducing the third term. Saito et al. \cite{saito2017asymmetric} adopted two classifiers to label the target set and made a constraint for the weights of two classifiers to make them different from each other. Zou et al. \cite{zou2018domain} introduced self-paced learning which generates pseudo labels from easy to hard to alleviate error accumulation of pseudo labels. Furthermore,  they utilized different confidence thresholds to select predicted target samples for various classes. Zou et al. \cite{zou2019confidence} introduced confidence regularization to prevent putting overconfident label belief in the wrong classes. On par with these methods that generate pseudo labels based on predictions, some methods generate pseudo labels in the feature space. Xie et al. \cite{xie2018learning} introduced feature centroids alignment after pseudo-labeling to DA. Chen et al. \cite{chen2019progressive} proposed a progressive feature alignment that takes advantage of intra-class distribution variance in pseudo-labeling for UDA problems.  Deng, Zheng, and Jiao \cite{DBLP:journals/corr/abs-1812-00893} introduced similarity-preserving constraint that can be implemented by minimizing the triplet loss with labeled source features and pseudo-labeled target features. Li et al. \cite{li2019locality} introduced label propagation to update pseudo labels and proposed landmark selection to re-weight the samples of the source and target domains. Wu et al. \cite{Wu2020IterativeRF} determined pseudo labels by integrating the predictions of multiple classifiers. Chen et al.  \cite{Chen2020DomainAB} improved the quality of pseudo labels by refining labels after each iteration. The proposed HCRPL is also based on pseudo-labeling and aims to improve the accuracy of pseudo labels by exploring the imbalance issue of performance among classes.

	\subsection{Semi-Supervised Domain Adaptation} 
	Since most DA methods focus on the unsupervised setting, SSDA has not been well studied. 	{ Several recent work \cite{Qin2020OppositeSL,Yang2020DeepCW,Li2019SemiSupervisedDA,Kim2020AttractPA,saito2019semi} show that SSDA can effectively boost the performance by adding merely few target labeled data (e.g. just one labeled image per class), suggesting that this setting may be more valuable in practical applications. For SSDA, the key to improving performance is learning target-discriminative representations \cite{saito2019semi}.} In Saito et al.'s study \cite{saito2019semi}, standard UDA methods \cite{ganin2014unsupervised,long2018conditional,saito2017adversarial} were shown to be empirically less effective in SSDA because they fail to learn discriminative class boundaries on the target domain. By optimizing a minimax loss on the conditional entropy of unlabeled data and a task loss, Saito et al. \cite{saito2019semi} reduced the distribution gap while learning discriminative features. Motiian et al. \cite{motiian2017unified}  exploited the Siamese architecture to learn an embedding subspace that is discriminative and where mapped visual domains are semantically aligned and yet maximally separated. Qin et al. \cite{qin2020opposite} proposed a framework consisting of a generator and two classifiers, where one is a source-based classifier and the other is a target-based classifier. The target-based classifier attempts to cluster the target features to improve intra-class density and enlarge inter-class divergence; the source-based classifier is designed to scatter the source features to enhance the smoothness of the decision boundary. Yan et al.\cite{Yan2018SemiSupervisedOT} proposed a semi-supervised entropic GromovWasserstein discrepancy approach to incorporate the supervision information when learning the optimal transport. {The proposed HCRPL can promote target-discriminative representations of the easy, normal, and hard classes. Hence, it also contributes to SSDA.}

		\section{Hard Class Problem}\label{sec_hard_class}
		In this section, we first design experiments to confirm the existence of the hard class problem and then emphasize the significance of alleviating it from two practical aspects.
	
	We set "Webcam" as the source domain and "Amazon" as the target domain and then adopt CBST to solve the domain shift problem between them. To investigate class-level performance, we choose precision, recall, and f1-score as metrics and report the results in Figure \ref{fig6}. Compared with training on source domain only, CBST enhances the performances on the majority of classes. We further observe the classes with lower precision, recall, or f1-score (e.g. 4-th and 28-th classes). These classes are so-called hard classes since low precision represents that massive predictions with high confidence are false, and low recall represents those correct predictions with high confidence are small. As a result, the hard class problem may further worsen the performance for these classes. For example, the precision, recall, and f1-score of the 28-th class decrease significantly.  
	
	Next, we emphasize two problems caused by the hard class problem. Firstly, it is noteworthy that the worst performance among all classes is more concerned rather than the average one in many applications. For instance,  in the application of transfer the knowledge of anomaly detection from the source domain to target one, we expect to detect the anomalies of every type, but the hard class problem will lead to the poor performance of hard classes. Secondly, the hard class problem also will have a negative effect on overall transfer performance, and our explanation is based on Ben-David et al's \cite{DBLP:journals/ml/Ben-DavidBCKPV10} theoretical analysis of DA.
	\begin{theorem}
		Let H be the hypothesis class. Given two different domains $\mathcal{S}$ and $\mathcal{T}$, we have
		\begin{equation}
			\forall{h}\in{\mathcal{H}},R_{\mathcal{T}}(h)\leq{R_{\mathcal{S}}(h)}+\frac{1}{2}d_{\mathcal{H}\Delta\mathcal{H}}(\mathcal{S},\mathcal{T})+C\label{eq:bound},
		\end{equation}
		where the expected error on target samples $R_{\mathcal{T}}(h)$ are bounded by three terms: (1) the expected error on source domain, $R_{\mathcal{S}}(h)$; (2) $d_{\mathcal{H}\Delta\mathcal{H}}(\mathcal{S},\mathcal{T})$ is the domain divergence measured by a discrepancy distance between the source domain distribution $\mathcal{S}$ and the target domain distribution $\mathcal{T}$ \emph{w.r.t.} a hypothesis set $\mathcal{H}$; and (3) the shared error of ideal joint hypothesis $C$. $C = \mathop{min}_{h \in {\mathcal{H}}}[\epsilon_{\mathcal{S}}(h, f_{\mathcal{S}}) + \epsilon_{\mathcal{T}}(h, f_{\mathcal{T}})]$, and $\epsilon_{\mathcal{S}}(h)$ is the expected error of $h$ on source domain.
	\end{theorem}
	
	In this study, we focus on the third term that is the shared error of ideal joint hypothesis $C$. According to triangle inequality for classification error \cite{DBLP:journals/ml/Ben-DavidBCKPV10}, that is, for any labeling functions $f_1, f_2$ and $f_3$, we have $\epsilon(f_1, f_2) \leq \epsilon(f_1, f_3) + \epsilon(f_2, f_3)$, we could have
	
	\begin{equation}
		\begin{aligned}
			C &= \mathop{min}_{h \in {\mathcal{H}}}\epsilon_{\mathcal{S}}(h, f_{\mathcal{S}}) + \epsilon_{\mathcal{T}}(h, f_{\mathcal{T}}) \\
			~ & \leq \mathop{min}_{h \in {\mathcal{H}}}\epsilon_{\mathcal{S}}(h, f_{\mathcal{S}}) + \epsilon_{\mathcal{T}}(h, f_{\mathcal{T}_{l}}) + \epsilon_{\mathcal{T}}(f_{\mathcal{T}_{l}}, f_{\mathcal{T}}),
		\end{aligned}
	\end{equation}
	$\epsilon_{\mathcal{S}}(h, f_{\mathcal{S}}) + \epsilon_{\mathcal{T}}(h, f_{\mathcal{T}_{l}})$ denotes the shared error of $h^*$ on source domain $\mathcal{S}$ and pseudo-labeled set $\mathcal{D}_{l}$ and is minimized by the training model with source domain $\mathcal{S}$ and pseudo-labeled set $\mathcal{D}_{l}$. $\epsilon_{\mathcal{T}}(f_{\mathcal{T}_{l}}, f_{\mathcal{T}})$ denotes false pseudo labels ratio. Overall, the expected error on the target domain is determined by the false pseudo labels ratio.
	
	Next, we link the hard class problem with the false pseudo labels ratio from the following two aspects. Firstly, few target samples are classified into hard classes, and the model cannot learn to target-discriminative representations for the hard classes.  Secondly, the hard class problem will cause error accumulation and increase the false pseudo labels ratio since pseudo-labeling may select false predictions with high confidence, and then these false predictions will misguide the classifier.
	
	\begin{algorithm}[htbp]
		\small
		\caption{Overall workflow for HCRPL}
		\label{algorithm1}
		\begin{algorithmic}[1]
			\Require rounds $Rs$, epochs $Es$, the source domain $\mathcal{D}_{s}$, the target domain $\mathcal{D}_{u}$, pre-trained network parameter $\theta$.
			\Ensure trained network parameter $\theta$. 
			\State Calculate the initial ensemble predictions $Z = \{z_i^u\}_{i=1}^{m_u}$ of the target domain $\mathcal{D}_{u}$ based on the pre-trained model.
			\State Let training set $D_{\text{tr}} = \mathcal{D}_{s}$.
			\State {Let pseudo-labeled set $\mathcal{D}_{l} = \emptyset$.}
			\For{$r = 1$ \textbf{to} $Rs$}
			\For{$e = 1$ \textbf{to} $Es$}
			\State Train network. \algorithmiccomment{Training phase}
			\State Update the ensemble predictions $Z$. \algorithmiccomment{Predicting phase}
			\EndFor
			\State {Select target samples with confident predictions and add them into pseudo-labeled set $\mathcal{D}_{l}$.}
			\algorithmiccomment{Selecting phase}
			\State Update training dataset $D_{\text{tr}} = \mathcal{D}_{s} \cup \mathcal{D}_{l}$.
			\EndFor
		\end{algorithmic}
	\end{algorithm}
	
	\section{Methods}
	\subsection{Preliminary}
	This section describes HCRPL based on the UDA setting. Under this setting, a source domain $\mathcal{D}_{s} = \{(x_i^s, y_i^s)\}_{i=1}^{m_s}$ and a target domain $\mathcal{D}_{u} = \{(x_i^u)\}_{i=1}^{m_u}$ are given. We define pseudo-labeled set as $\mathcal{D}_{l} = \{(x_i^u, \hat{y}^u_i\}^{m_t}_{i=1}$. Specifically, $y_i^s$ and $\hat{y}^u_i$ are one-hot vectors. Meanwhile, we assume that source and target domains contain same object classes, and we consider $C$ classes. Under the SSDA setting, the union of source domain and labeled target set is regarded as the new source domain while the unlabeled target set is regarded as the new target domain.
	
\begin{figure}[!t]
		\centering
		\includegraphics[width=\columnwidth]{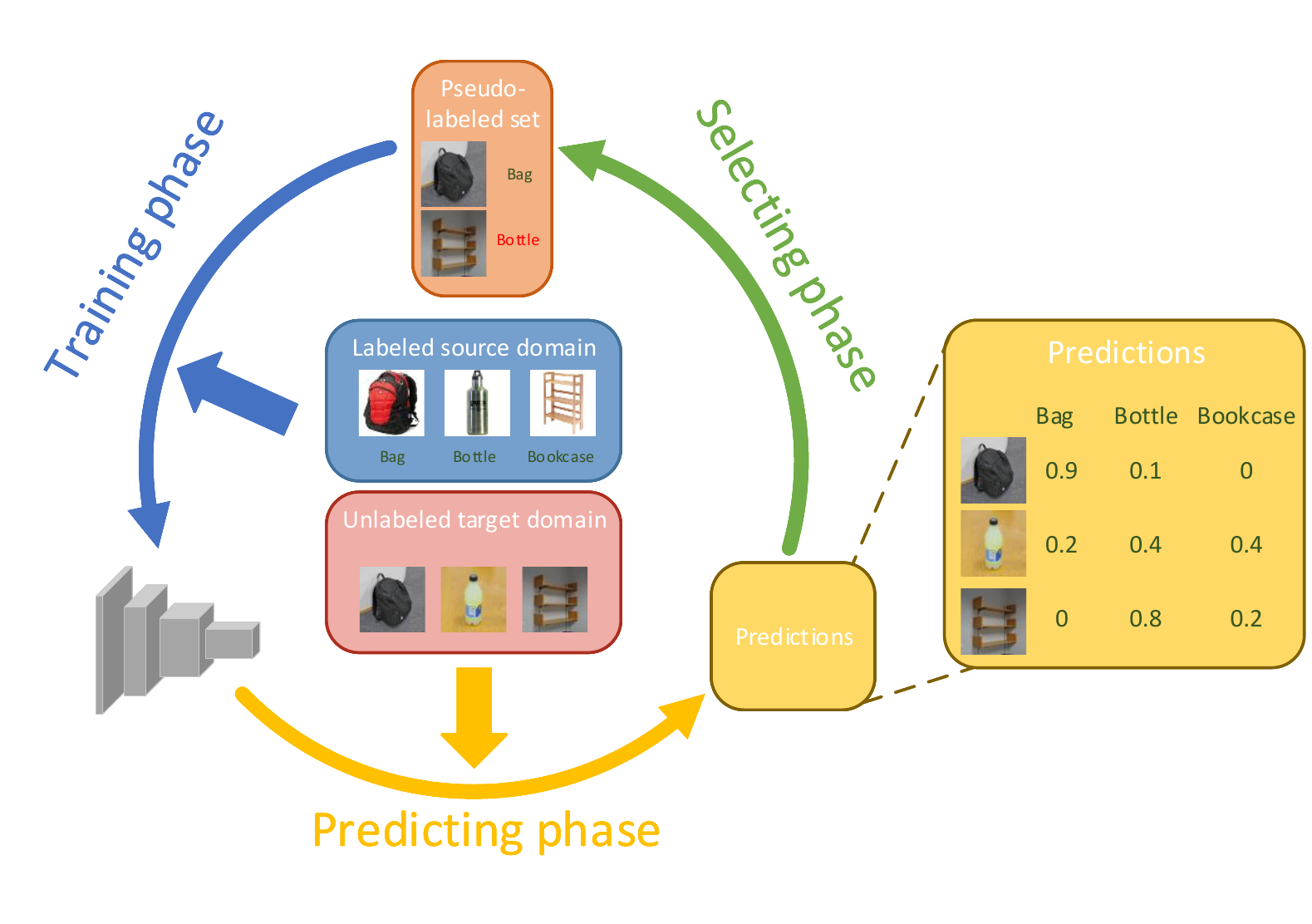}
		\caption{The overall framework of the proposed HCRPL, which is mainly composed of training, predicting, and selecting phases. The whole framework undergoes these three phases alternatively. 
		}\label{fig2}
	\end{figure}

	The proposed HCRPL belongs to pseudo-labeling, but it is different from the standard pseudo-labeling where target samples are predicted after every epoch but not every round. To describe the proposed HCRPL more clearly, the labeling phase is divided into predicting and selecting phases. The overall training process is given in Algorithm \ref{algorithm1} and Figure \ref{fig2}. It mainly includes three phases as follows:
	\begin{itemize}
		\item [1)]
		Training phase: training network with training set $\mathcal{D}_{tr}$.
		\item [2)]
		Predicting phase: generating ensemble predictions $Z$ of target samples $\mathcal{D}_{u}$.
		\item [3)]
		Selecting phase: selecting target samples with confident predictions as the pseudo-labeled set $\mathcal{D}_{l}$.
	\end{itemize}
	
	We define going through the process from training the network to updating $\mathcal{D}_{tr}$ once as one round. The proposed APC, TE, and SE are included in the predicting phase. The overall training process along with APC, TE, and SE are introduced below.

	\begin{figure*}
		\centering
		\includegraphics[width=6.9in]{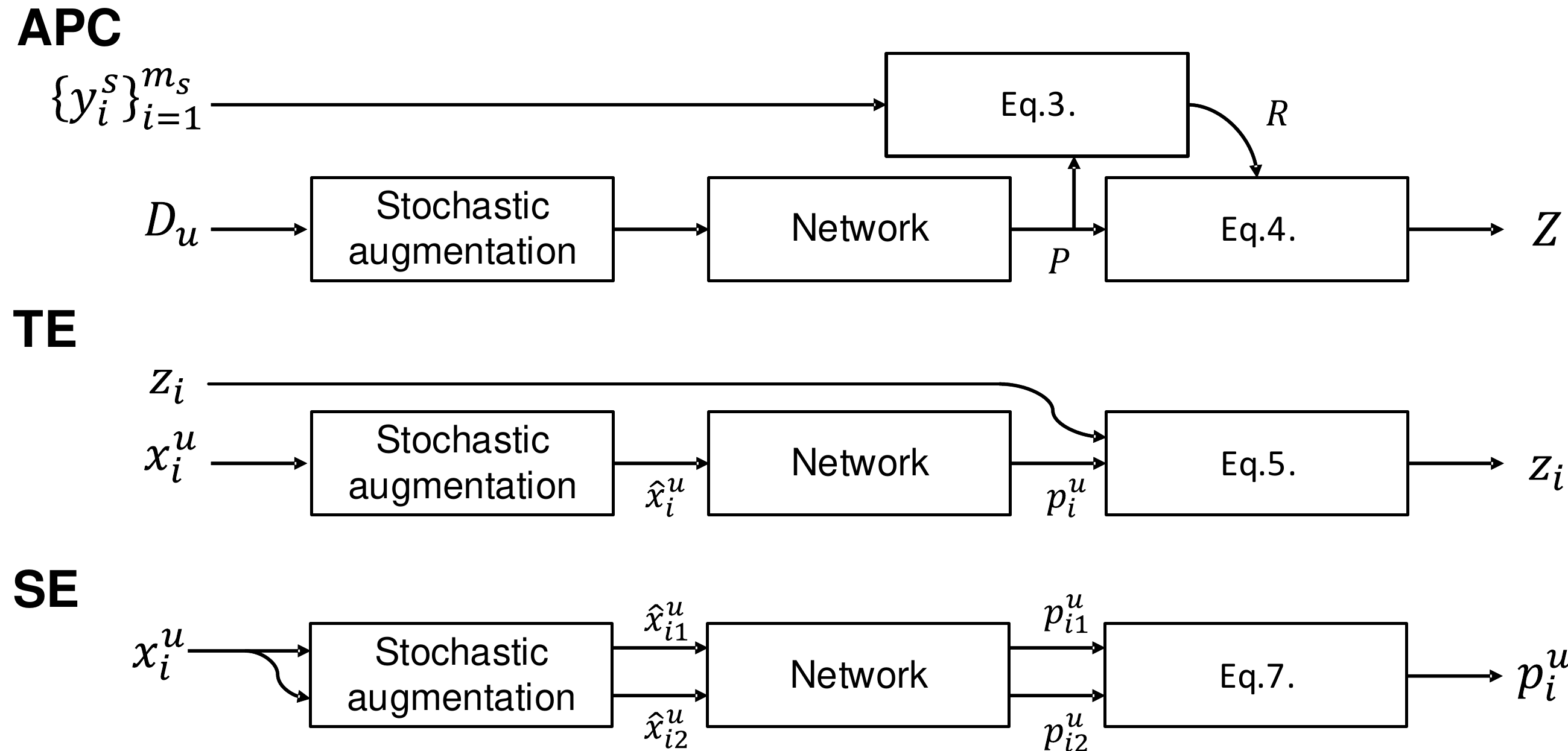}
		\caption{Structure of the training pass in HCRPL. Top: APC. Middle: TE. Bottom: SE. the details of Eq.3., Eq.4., Eq.5. and Eq.7. are show in  Equal \ref{eq1}, \ref{eq2}, \ref{eq3} and \ref{eq5}.
		}\label{fig3}
	\end{figure*}
	
	\subsection{Adaptive Prediction Calibration}
	Adaptive prediction calibration is the core component of our approach. In the hard class problem, target samples are difficult to be classified into hard ones, which leads to lower predictive class proportions for hard classes. One way to alleviate this problem is magnifying the predictive proportions of the hard classes while keeping the normal ones and suppressing the easy ones. To control predictive class proportions in a reasonable interval, we try to eliminate the mismatch between predictive and true class proportions for the target domain. However, the latter is always unknown in practice. Here, we replace it with the class proportion of the source domain, which is reasonable due to the following two aspects. First, the label proportions of the source and target domains always are close, which is valid in many practical applications. Second, we demonstrate that the minor disagreement of label proportions slightly affects transfer performance by experiments, the details of which are shown in Section \ref{pcp}.
	
	The detailed process of APC is shown in Figure \ref{fig3} top. The target domain $\mathcal{D}_{u}$ is first fed into the trained model to obtain predictions $P = \{p_i^u\}_{i=1}^{m_u}$. Then, we define a ratio $R$ of class distribution of the source domain $q(y)$ to predictive class distribution $p(y)$ as follows
	\begin{equation}\label{eq1}
		R = q(y) \oslash p(y),
	\end{equation}
	where $q(y)=\frac{1}{m_s}\sum_{i=1}^{m_s}y_i^s$, $p(y)=\frac{1}{m_u}\sum_{i=1}^{m_u}p_i^u$, $\oslash$ means element-wise division, and $R$ is a $C$-dimensional vector with $i$-th dimension being the difficulty degree belonging to $i$-th class. Finally, we calibrate predictions $P$ by
	\begin{equation}\label{eq2}
		P \leftarrow \{\text{Normalization}(R \odot p_i^u)\}_{i=1}^{m_u},
	\end{equation}
	where $\text{Normalization}(x) = \frac{x}{\sum_i x_i}$ and $\odot$ means element-wise multiplication. Intuitively, we calibrate $P$ by $R$. For a certain class $c$, if the predictive probability of class $c$ is small, which means that class $c$ is the hard class,  the APC will increase the probabilities of classifying target samples into class $c$.

	\subsection{Temporal-Ensembling and Self-Ensembling}
	To ensure the reliability of predictions, we further introduce two ensembling methods, TE and SE. For TE, integrating predictions of multiple classifiers is considerable to obtain consistent predictions. Different from ATDA \cite{saito2017asymmetric} that constructed two classifiers with multiview loss, we adopt the temporal-ensembling based method, which views the trained model after each epoch as a classifier. Hence, there is no need to construct multiple classifiers, which saves the number of parameters and avoids the multiview loss. As shown in Figure~\ref{fig3} middle, we evaluate the model on the target domain after every epoch and update ensemble predictions $z_i$ by Exponential Moving Average (EMA). The EMA can be formulated as
	\begin{equation} \label{eq3}
		z_i \leftarrow \alpha z_i + (1-\alpha)p_i^u.
	\end{equation}
	The EMA can memorize all predictions and place a greater weight on the recent ones. $\alpha$ is the EMA momentum, and recent predictions will have a higher proportion with a lower $\alpha$. If $\alpha =0$, ensemble predictions $z_i$ are equal to current ones $p_i^u$. Specifically, ensemble predictions $z_i$ are used to select pseudo-labeled samples in the selecting phase.
	
	For SE, we integrate the predictions of two different augmentations. As shown in Figure \ref{fig3} bottom, we feed target samples into the trained model twice with different stochastic augmentations and obtain predictions $p_{i1}^u$ and $p_{i2}^u$. Then, the average predictions $\frac{P_1+P_2}{2}$ are calculated. To obtain lower entropy predictions, we perform an additional step, named Sharpening \cite{berthelot2019mixmatch,berthelot2019remixmatch}. It is defined as:
	\begin{equation}
		\text{Sharpen}(p, T) = \text{Normalization}(p^{1/T}),
	\end{equation}
	where $T$ is named as the sharpening temperature. When $T$ is smaller, predictions with lower entropy are obtained. Finally, we obtain predictions $P$ by
	\begin{equation}\label{eq5}
		P = \text{Sharpen}(\frac{P_1+P_2}{2}, T).
	\end{equation}
	
	TE and SE are somewhat similar to $\Pi$-model \cite{laine2016temporal} and Mean Teacher \cite{DBLP:conf/iclr/FrenchMF18,DBLP:conf/iclr/TarvainenV17}. These methods took the predictive difference of different epochs and augmentations as a regularization term to constrain the model. Here, we consider that the distributional mismatch and the deviation magnification result in the vulnerability of predictions. Hence, it is adequately necessary to introduce TE and SE to stabilize the predictions.
	
	\begin{algorithm}
		\small
		\caption{Details of the prediction process}
		\label{algorithm2}
		\begin{algorithmic}[1]
			\Require ensemble predictions shadow value $Z$, prior class proportion $q(y)$, the target domain $\mathcal{D}_{u}$. network parameter $\theta$,  EMA momentum $\alpha$.
			\Ensure updated ensemble predictions shadow value $Z$.
			\State Let $P_1=\emptyset, P_2=\emptyset$.
			\For {$i = 1$ \textbf{to} $m_u$}
			\State $\hat{x}_{i1}^u =$ Augment $(x_i^u)$
			\State $\hat{x}_{i2}^u =$ Augment $(x_i^u)$
			\State $P_1 \leftarrow P_1 + p_{i1}^u$ 
			\algorithmiccomment{$p_{i1}^u=p(y|\hat{x}_{i1}^u, \theta)$}
			\State $P_2 \leftarrow P_2 + p_{i2}^u$ 
			\algorithmiccomment{$p_{i2}^u=p(y|\hat{x}_{i2}^u, \theta)$}
			\EndFor
			\State $p(y) = \frac{1}{2m_u}(\sum_{i=1}^{m_u}(p_{i1}^u + p_{i2}^u))$
			\State $R = q(y) \oslash p(y)$
			\State $P_1 \leftarrow \{\text{Normalization}(R \otimes p_{i1}^u)\}_{i=1}^{m_u}$
			\State $P_2 \leftarrow \{\text{Normalization}(R \otimes p_{i2}^u)\}_{i=1}^{m_u}$
			\State $P = \text{Sharpen}(\frac{P_1+P_2}{2}, T)$
			\State $Z \leftarrow \alpha Z + (1-\alpha)P$
			\State \textbf{return} $Z$
		\end{algorithmic}
	\end{algorithm}
	
	\subsection{Overall Training Process}
	The details of training, predicting, and selecting phases are described below.
	
	In the training phase, the network is trained with the training dataset $D_{\text{tr}}$. In the first round, we view the source domain $\mathcal{D}_{s}$ as training dataset $D_{\text{tr}}$. Subsequently, the training dataset is $D_{\text{tr}}$ updated  with the union of the pseudo-labeled target set $\mathcal{D}_{l}$ and source domain $\mathcal{D}_{s}$. The training objective is defined as
	
	\begin{equation}
		\begin{aligned}
			L = \frac{1}{m_s+ m_t} (\sum_{i=1}^{m_s} \text{H}(y_i^s, p_i^s)+
			\sum_{i=1}^{m_t} \text{H}(\hat y_i^u, p_i^u)),
		\end{aligned}
	\end{equation}
	where $\text{H}(p,q)$ is a standard cross-entropy loss function. With the number of pseudo-labeled target samples increasing, the network learns more target-discriminative representations and gradually enhances the transfer performance on the target domain.

	In the predicting phase, we aim to achieve accurate and robust predictions, especially for hard classes. We propose the APC, TE, and SE to adjust the predictions of target samples.  The overall workflow pseudo-code for the predicting phase is given in Algorithm \ref{algorithm2}. The predicting phase can be split into five parts: Firstly, the target set is augmented twice and the corresponding predictions $P_1, P_2$  are obtained (Line 2-7). Secondly, the difficulty ratio $R$ of prior class proportion $q(y)$ to predictive class distribution $p(y)$  is calculated (Line 8-9). Thirdly, the APC is applied to calibrate predictions $P_1, P_2$ (Line 10-11). Fourthly, average predictions $\frac{P_1+P_2}{2}$  are calculated (Line 12). Finally, ensemble predictions $Z$ are updated by EMA (Line 13).
	
	In the selecting phase, we select the predictions with higher confidence as pseudo labels. CBST \cite{zou2018domain} considers that different classes should have different confidence thresholds and dynamically adjusts thresholds to generate pseudo labels from easy to hard. The class-balanced self-training solver can be formulated as
	\begin{equation}
		\hat{y}_{ic}^{u} = \left\{
		\begin{aligned}
			1&, \text{if}\ c = \mathop{\text{argmax}}_c \frac{z_{ic}^u}{\text{exp}(-k_c)},\\
			\quad&\frac{z_{ic}^u}{\text{exp}(-k_c)} >1.\\
			0&, \text{otherwise}.
		\end{aligned}
		\right.
	\end{equation}
	Here, $z_i^u$ means the ensemble predictions of $i$-th sample. $z_{ic}^u$ means the $c$-th class probability for $z_i^u$. $k_c$ for each class $c$ is determined by a single portion parameter $p$ which controls the number of selected samples. Practically, $p$ gradually increases to select more pseudo-labeled samples. For a detailed algorithm, we recommend reading Algorithm 2 in Zou et al's paper \cite{zou2018domain}.

	\begin{figure*}
		\centering
		\includegraphics[width=2.1\columnwidth]{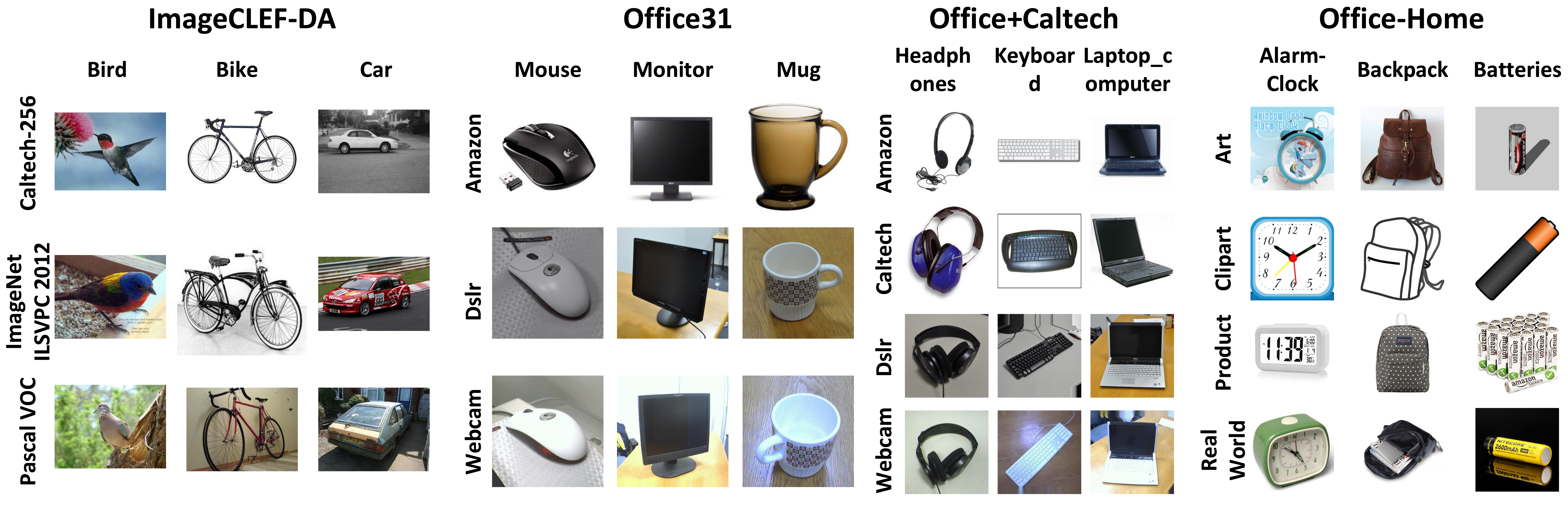}
		\caption{Example images in ImageCLEF-DA, Office-31, Office+Caltech, and Office-Home}
		\label{fig4}
	\end{figure*}

	\section{Experiments}
	\subsection{Datasets}

	\subsubsection{ImageCLEF-DA} ImageCLEF-DA$\footnote{http://imageclef.org/2014/adaptation}$ is a benchmark dataset for ImageCLEF 2014 DA challenges. Three domains, including Caltech-256 (C), ImageNet ILSVRC 2012 (I), and Pascal VOC (P), share 12 categories. Each domain contains 600 images and 50 images for each category.
	
	\subsubsection{Office-31} Office-31 \cite{saenko2010adapting} is a standard benchmark for DA tasks. The dataset contains $4110$ images of $31$ categories collected from three domains: Amazon (A), Webcam (W), and Dslr (D). Under the SSDA setting, we followed the settings used by Saito et al. \cite{saito2019semi} and then evaluated the proposed method on the task between W $\rightarrow$ A and D $\rightarrow$ A.
	
	{\subsubsection{Office+Caltech}
		Office+Caltech \cite{gong2012geodesic} is a common benchmark that consists of four domains: Caltech, Amazon, Webcam, and Dslr. These four domains share images from ten categories. There are 2533 images in total.}
	
	\subsubsection{Office-Home} Office-Home \cite{venkateswara2017deep} is a more challenging DA dataset compared to Office-31. It consists of 15500 images from 65 categories and is organized into four domains: Art (Ar), Clipart (Cl), Product (Pr), and Real-world (Rw).
	
	We show examples of four datasets in Figure \ref{fig4}. We can see different classes have different domain shifts. For example, in Office-Home, the examples of class 'Alarm-Clock' from different domains are similar to each other whereas the examples of class 'Backpack' vary a lot.

	\subsection{Baselines}
	\subsubsection{Unsupervised Domain Adaptation} 
	
	{For Office+Caltech, we set AlexNet \cite{krizhevsky2012imagenet} as backbone and compare our approach with traditional and deep methods, such as SA \cite{6751479}, GFK \cite{gong2012geodesic}, TCA \cite{pan2010domain}, SCA \cite{Ghifary2017ScatterCA}, LPJT \cite{Jingjing2019LocalityPJ}, KJDIP-rbf \cite{Chen2020DomainAB}, FSDA \cite{Sun2019InformativeFS}, MMD-CORAL \cite{Rahman2020OnMD}, and GKE \cite{Wu2020GeometricKE}.} For the remaining datasets,  we evaluated HCRPL with $\textbf{ResNet-50}$ \cite{he2016deep}. We compared the proposed HCRPL with the latest methods, including Reverse Gradient ($\textbf{RevGrad}$) \cite{ganin2014unsupervised}, Joint Adaptation Networks ($\textbf{JAN}$) \cite{long2017deep}, Class-Balanced Self-Training ($\textbf{CBST}$) \cite{zou2018domain}, Confidence Regularized Self-Training ($\textbf{CRST}$) \cite{zou2019confidence}, $\textbf{CDAN}$ \cite{long2018conditional},  $\textbf{SAFN}$\cite{xu2019larger}, Domain-Symmetric Networks ($\textbf{SymNets}$) \cite{zhang2019domain}, Domain adversarial neural network ($\textbf{DANN}$) \cite{ganin2016domain}, Discriminative Adversarial Domain Adaptation ($\textbf{DADA}$)  \cite{tang2019discriminative}, $\textbf{MDD}$ \cite{zhang2019bridging}, Maximum Classifier Discrepancy ($\textbf{MCD}$) \cite{saito2018maximum} and Cycle-consistent Conditional Adversarial Transfer Networks ($\textbf{3CATN}$) \cite{li2019cycle}.
	
	\subsubsection{Semi-Supervised Domain Adaptation} In the SSDA experiments, we evaluated the proposed model with $\textbf{AlexNet}$ \cite{krizhevsky2012imagenet} and $\textbf{VGG16}$ \cite{simonyan2014very}. The proposed model was compared with Domain adversarial neural network ($\textbf{DANN}$) \cite{ganin2016domain}, $\textbf{CDAN}$ \cite{long2018conditional}, $\textbf{ENT}$ \cite{grandvalet2005semi}, Adversarial dropout regularization ($\textbf{ADR}$) \cite{saito2017adversarial}, and Minimax Entropy ($\textbf{MME}$) \cite{saito2019semi}. 
	
		\begin{center}
		\begin{table*}
			\small
			\centering
			\caption{ResNet50-based approaches on Office-Home under the UDA setting (\%)}
			\begin{tabular*}{\hsize}{@{}@{\extracolsep{\fill}}cccccccccccccc@{}}
				\toprule
				\multirow{2}{*}{Method} & Ar & Ar & Ar & Cl & Cl & Cl & Pr & Pr & Pr & Rw & Rw & Rw & \multirow{2}{*}{Avg}\\
				~ & Cl & Pr & Rw & Ar & Pr & Rw & Ar & Cl & Rw & Ar & Cl & Pr & ~\\
				\midrule
				ResNet50 \cite{he2016deep} & 34.9 & 50.0 & 58.0 & 37.4 & 41.9 & 46.2
				& 38.5 & 31.2 & 60.4 & 53.9 & 41.2 & 59.9 & 46.1\\
				RevGrad \cite{ganin2014unsupervised} & 45.6 & 59.3 & 70.1 & 47.0 & 58.5 & 60.9
				& 46.1 & 43.7 & 68.5 & 63.2 & 51.8 & 76.8 & 57.6\\
				CBST \cite{zou2018domain}  & 51.4 & 74.1 & 78.9 & 56.3 & 72.2 & 73.4
				& 54.4 & 41.6 & 78.8 & 66.0 & 48.3 & 81.0 & 64.7\\
				CDAN+E \cite{long2018conditional}  & 50.7 & 70.6 & 76.0 & 57.6 & 70.0 & 70.0
				& 57.4 & 50.9 & 77.3 & 70.9 & 56.7 & 81.6 & 65.8\\
				AADA+CCN \cite{yang2020mind} & 54.0 & 71.3 & 77.5 & 60.8 & 70.8 & 71.2
				& 59.1 & 51.8 & 76.9 & 71.0 & 57.4 & 81.8 & 67.0\\ 	
				SAFN \cite{xu2019larger}  & 52.0 & 71.7 & 76.3 & 64.2 & 69.9 & 71.9
				& 63.7 & 51.4 & 77.1 & 70.9 & 57.1 & 81.5 & 67.3\\ 	
				SymNets \cite{zhang2019domain}  & 47.7 & 72.9 & 78.5 & 64.2 & 71.3 & 74.2
				& \textbf{64.2} & 48.8 & 79.5 & \textbf{74.5} & 52.6 & 82.7 & 67.6\\
				ATM \cite{li2020maximum}  & 52.4 & 72.6 & 78.0 & 61.1 & 72.0 & 72.6
				& 59.5 & 52.0 & 79.1 & 73.3 & 58.9 & 83.4 & 67.9\\
				MDD \cite{zhang2019bridging}  & 54.9 & 73.7 & 77.8 & 60.0 & 71.4 & 71.8
				& 61.2 & \textbf{53.6} & 78.1 & 72.5 & 60.2 & 82.3 & 68.1\\
				GSDA \cite{hu2020unsupervised}  & \textbf{61.3} & 76.1 & 79.4 & 65.4 & 73.3 & 74.3
				& 65.0 & 53.2 & 80.0 & 72.2 & \textbf{60.6} & 83.1 & 70.3\\
				\midrule
				HCRPL (proposed)  & 59.5 & \textbf{76.8} & \textbf{80.8} & \textbf{67.2} & \textbf{76.7} & \textbf{78.9}
				& 63.2 & 53.1 & \textbf{81.2} & 72.3 & 57.2 & \textbf{84.4} & \textbf{70.9}\\
				\bottomrule
			\end{tabular*}
			\label{table1}
		\end{table*}
	\end{center}

	\begin{center}
		\begin{table*}
			\small
			\centering
			\caption{ResNet50-based approaches on Office-31  under the UDA setting (\%)}
			\begin{tabular*}{\hsize}{@{}@{\extracolsep{\fill}}cccccccc@{}}
				\toprule
				Method & A $\rightarrow$ W & D $\rightarrow$ W & W $\rightarrow$ D & A $\rightarrow$ D & D $\rightarrow$ A & W $\rightarrow$ A & Avg\\
				\midrule
				ResNet50 \cite{he2016deep} & 68.4$\pm$0.2 & 96.7$\pm$0.2 & 99.3$\pm$0.1 & 68.9$\pm$0.2 & 62.5$\pm$0.3 & 60.7$\pm$0.3 & 76.1\\
				DANN \cite{ganin2016domain}  &82.0$\pm$0.4 & 96.9$\pm$0.2 & 99.1$\pm$0.1 & 79.7$\pm$0.4 & 68.2$\pm$0.4 & 67.4$\pm$0.5 & 82.2\\
				CBST \cite{zou2018domain}  &87.8$\pm$0.8 & 98.5$\pm$0.1 & \textbf{100.0}$\pm$0.0 & 86.5$\pm$1.0 & 71.2$\pm$0.4 & 70.9$\pm$0.7 & 85.8\\
				MCD \cite{saito2018maximum}  & 89.6$\pm$0.2 & 98.5$\pm$0.1 & \textbf{100.0}$\pm$0.0 & 91.3$\pm$0.2 & 69.6$\pm$0.3 & 70.8$\pm$0.3 & 86.6\\
				CRST \cite{zou2019confidence}  &89.4$\pm$0.7 & 98.9$\pm$0.4 & \textbf{100.0}$\pm$0.0 & 88.7$\pm$0.8 & 72.6$\pm$0.7 & 70.9$\pm$0.5 & 86.8\\
				SAFN+ENT \cite{xu2019larger}  & 90.1$\pm$0.8 & 98.6$\pm$0.2 & 99.8$\pm$0.0 & 90.7$\pm$0.5 & 73.0$\pm$0.2 & 70.2$\pm$0.3 & 87.1\\
				CDAN+E \cite{long2018conditional}  & 94.1$\pm$0.1 & 98.6$\pm$0.1 & \textbf{100.0}$\pm$0.0 & 92.9$\pm$0.2 & 71.0$\pm$0.3 & 69.3$\pm$0.3 & 87.7\\
				SymNets \cite{zhang2019domain}  & 90.8$\pm$0.1 & 98.8$\pm$0.3 & \textbf{100.0}$\pm$0.0 & 93.9$\pm$0.5 & 74.6$\pm$0.6 & 72.5$\pm$0.5 & 88.4\\
				MDD \cite{zhang2019bridging}  & 94.5$\pm$0.3 & 98.4$\pm$0.1 & \textbf{100.0}$\pm$0.0 & 93.5$\pm$0.2 & 74.6$\pm$0.3 & 72.2$\pm$0.1 & 88.9\\
				3CATN \cite{li2019cycle} & 95.3$\pm$0.3 & \textbf{99.3}$\pm$0.5 & \textbf{100.0}$\pm$0.0 & 94.1$\pm$0.3 & 73.1$\pm$0.2 & 71.5$\pm$0.6 & 88.9\\

				DADA \cite{tang2019discriminative}  &92.3$\pm$0.1 & 99.2$\pm$0.1 & \textbf{100.0}$\pm$0.0 & 93.9$\pm$0.2 & 74.4$\pm$0.1 & 74.2$\pm$0.1 & 89.0\\
				GSDA \cite{hu2020unsupervised} & 95.7 & 99.1 & \textbf{100.0} & 94.8 & 73.5 & 74.9 & 89.7\\
				ATM \cite{li2020maximum} & 95.7$\pm$0.2 & \textbf{99.3}$\pm$0.1 & \textbf{100.0} $\pm$ 0.0 & \textbf{96.4} $\pm$ 0.2 & 74.1 $\pm$ 0.2 & 73.5 $\pm$ 0.3 & 89.8\\
				\midrule
				HCRPL (proposed) & \textbf{95.9}$\pm$0.2 & 98.7$\pm$0.1 & \textbf{100.0}$\pm$0.0 & 94.3$\pm$0.2 & \textbf{75.0}$\pm$0.4 & \textbf{75.4}$\pm$0.4 & \textbf{89.9}\\
				\bottomrule
			\end{tabular*}
			\label{table2}
		\end{table*}
	\end{center}

	\begin{center}
		\begin{table*}
			\scriptsize
			\centering
			\caption{Shallow and deep approaches on Office+Caltech under the UDA setting (\%)}
			\begin{tabular*}{\hsize}{@{}@{\extracolsep{\fill}}cccccccc|cccc@{}}
				\toprule
				\multirow{2}{*}{S $\rightarrow$ T} & SA \cite{6751479} & GFK \cite{gong2012geodesic} & TCA \cite{pan2010domain} & SCA \cite{Ghifary2017ScatterCA} & LPJT \cite{Jingjing2019LocalityPJ} & FSDA \cite{Sun2019InformativeFS} & KJDIP-rbf \cite{Chen2020DomainAB} & AlexNet \cite{krizhevsky2012imagenet} & MMD-CORAL \cite{Rahman2020OnMD} & GKE \cite{Wu2020GeometricKE} & HCRPL \\
				~ & \multicolumn{7}{c}{Shallow} & \multicolumn{4}{c}{Deep} \\
				\midrule
				A$\rightarrow$ C & 80.1 & 76.9 & 74.2 & 78.8 & 85.4 & 88.3 & 85.8 & 84.6 & \textbf{89.1} & 88.4 & \textbf{89.1}\\
				A$\rightarrow$ D & 78.3 & 79.6 & 78.3 & 85.4 & -    & 87.9 & 87.9 & 88.5 & 96.6 & \textbf{99.7} & 95.8\\
				A$\rightarrow$ W & 68.8 & 68.5 & 71.9 & 75.9 & 92.2 & 82.7 & 91.2 & 83.1 & 95.7 & \textbf{97.6} & 95.9\\
				C$\rightarrow$ A & 89.5 & 88.4 & 89.3 & 89.5 & 92.1 & 92.8 & 92.4 & 91.8 & 93.6 & 93.5 & \textbf{94.0}\\
				C$\rightarrow$ D & 83.4 & 84.6 & 83.4 & 87.9 & -    & 91.1 & 90.4 & 89.0 & 93.4 & 94.3 & \textbf{98.3}\\
				C$\rightarrow$ W & 75.9 & 80.7 & 80.0 & 85.4 & 92.7 & 88.8 & 89.5 & 83.1 & 95.2 & 98.3 & \textbf{98.8}\\
				D$\rightarrow$ A & 82.7 & 85.8 & 88.2 & 90.0 & -    & 89.3 & 89.4 & 89.3 & 94.7 & 93.5 & \textbf{94.8}\\
				D$\rightarrow$ C & 75.7 & 74.1 & 73.5 & 78.1 & -    & 80.0 & 78.5 & 80.9 & 84.7 & 83.8 & \textbf{89.2}\\
				D$\rightarrow$ W & 99.3 & 98.6 & 97.3 & 98.6 & -    & 98.0 & 97.6 & 97.7 & 99.4 & \textbf{99.7} & 99.6\\
				W$\rightarrow$ A & 77.8 & 75.3 & 80.0 & 86.1 & 92.3 & 87.6 & 92.1 & 83.8 & \textbf{94.8} & 94.4 & 94.6\\
				W$\rightarrow$ C & 74.9 & 74.8 & 72.6 & 74.8 & 86.0 & 80.1 & 83.5 & 77.7 & 86.5 & 88.9 & \textbf{88.9}\\
				W$\rightarrow$ D & \textbf{100.0} & \textbf{100.0} & \textbf{100.0} & \textbf{100.0} & -   & 99.4 & 96.8 & \textbf{100.0}& \textbf{100.0}& \textbf{100.0}& \textbf{100.0}\\
				Avg              & 82.2 & 82.4 & 82.4 & 85.9 & -    & 88.8 & 89.6 & 87.5 & 93.6 & 94.3 & \textbf{94.9}\\
				\bottomrule
			\end{tabular*}
			\label{tab4}
		\end{table*}
	\end{center}

	\subsection{Implementation Details}
	The proposed HCRPL was implemented with PyTorch$\footnote{https://pytorch.org/}$. For a fair comparison, our backbone network was the same as the competitive methods. For AlexNet, we replace the last full-connected layer with a randomly initialized bottleneck layer which consists of two fully-connected layers: $4096 \rightarrow C$. For VGG and ResNet, we replace the last full-connected layer with a randomly initialized $C$-way classifier layer. It was pre-trained on ImageNet and then fine-tuned using SGD with a weight decay of $5 \times 10^{-4}$, the momentum of $0.9$, and the batch size of $32$. Likewise, we used horizontal-flipping and random-cropping based data augmentation for all the training data. For the pseudo-labeling setting, we set the total number of pseudo-labeling rounds to be 30, each with 20 epochs. In the $r$-th round, we choose the top $p\%$ of the highest confidence target samples within each category, and $p=min(r*5+10, 90)$. In the first round, the network was trained with labeled data (the source domain in UDA; the source domain and labeled target data in SSDA) with a learning rate of $5 \times 10^{-5} $ or $1 \times 10^{-4}$. Furthermore, it was then retrained in the subsequent rounds with a learning rate of $1.5 \times 10^{-5}$. We set the EMA momentum $\alpha$ to 0.95. Under the SSDA setting, we added a few-shot module to AlexNet and VGG16 like Saito et al. \cite{saito2019semi} for better comparison with MME.

	\begin{center}
	\begin{table*}
		\small
		\centering
		\caption{Results on ImageCLEF-DA dataset under the UDA setting(\%)}	
		\begin{tabular*}{\hsize}{@{}@{\extracolsep{\fill}}cccccccc@{}}
			\toprule
			Method & I $\rightarrow$ P & P $\rightarrow$ I & I $\rightarrow$ C & C $\rightarrow$ I & C $\rightarrow$ P & P $\rightarrow$ C & Avg\\
			\midrule
			ResNet50 \cite{he2016deep} & 74.8 & 83.9 & 91.5 & 78.0 & 65.5 & 91.2 & 80.7\\
			DANN \cite{ganin2016domain} & 75.0 & 86.0 & 96.2 & 87.0 & 74.3 & 91.5 & 85.0 \\
			MCD \cite{saito2018maximum} & 77.3 & 89.2 & 92.7 & 88.2 & 71.0 & 92.3 & 85.1 \\
			JAN \cite{long2017deep} & 76.8 & 88.0 & 94.7 & 89.5 & 74.2 & 91.7 & 85.8 \\
			CBST \cite{zou2018domain} & 77.8 & 91.7 & 96.2 & 91.1 & 75.0 & 93.9 & 87.6 \\
			CDAN+E \cite{long2018conditional} & 77.7 & 90.7 & \textbf{97.7} & 91.3 & 74.2 & 94.3 & 87.7 \\
			SAFN \cite{xu2019larger} & 78.0 & 91.7 & 96.2 & 91.1 & 77.0 & 94.7 & 88.1 \\
			AADA+CCN \cite{yang2020mind} & \textbf{79.2} & 92.5 & 96.2 & 91.4 & 76.1 & 94.7 & 88.4 \\
			\midrule
			HCRPL  & 78.2 & \textbf{92.9} & 96.6 & \textbf{92.3} & \textbf{77.5} & \textbf{95.8} & \textbf{88.9} \\
			\bottomrule
		\end{tabular*}
		\label{table5}
	\end{table*}
\end{center}

	\subsection{Results} \label{5.4}
	\subsubsection{Unsupervised domain adaptation} Transfer performances on Office-Home, Office-31, Office+Caltech, and  ImageCLEF-DA datasets under the UDA setting are shown in Table \ref{table1}, \ref{table2}, \ref{tab4}, and \ref{table5}, respectively. For Office-Home, the result is shown in Table \ref{table1}, and HCRPL outperforms the best performance by $2.8\%$ on average and achieves state-of-the-art performance on most tasks. It should be noted that the proposed framework has a larger improvement on the transfer tasks with larger domain shifts. For example, tasks Ar $\rightarrow$ Cl and  Cl $\rightarrow$ Ar have poor generalization ability on the target domain, and HCRPL has an improvement over MDD by $5.4\%$ on Ar $\rightarrow$ Cl and $7.2\%$ on Cl $\rightarrow$ Ar. For Office-31, we report the averages and standard deviations of evaluation results over 3 runs. HCRPL outperforms all the other methods on $A \rightarrow W$, $A \rightarrow D$, $D \rightarrow A$, $D \rightarrow A$, and the average of all sub-tasks. {For Office+Caltech, except for AlexNet based deep methods, our approach also compares with the shallow methods, which replace the image input with features extracted by Decaf \cite{Donahue2014DeCAFAD} or VGG \cite{simonyan2014very}.  In general, deep methods are better than shallow ones, and our approach further outperforms some existing deep methods.}  For ImageCLEF-DA, HCRPL outperforms all the other methods on $I \rightarrow P$, $P \rightarrow I$, $C \rightarrow I$, $C \rightarrow P$, $P \rightarrow C$, and the average of all sub-tasks.

	\begin{figure*}
		\centering
		\subfigure[Source-only]{
			\label{fig81}
			\includegraphics[width=2.2in]{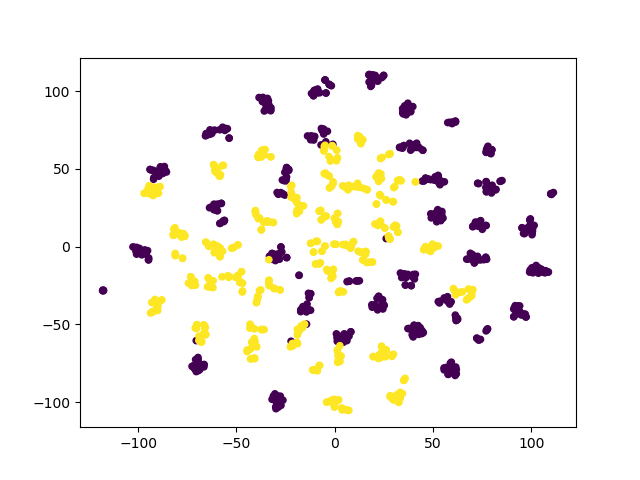}}
		\subfigure[CBST]{
			\label{fig82}
			\includegraphics[width=2.2in]{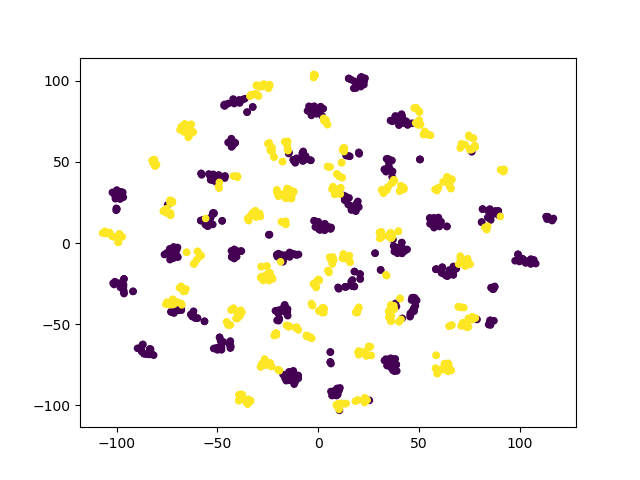}}
		\subfigure[HCRPL]{
			\label{fig83}
			\includegraphics[width=2.2in]{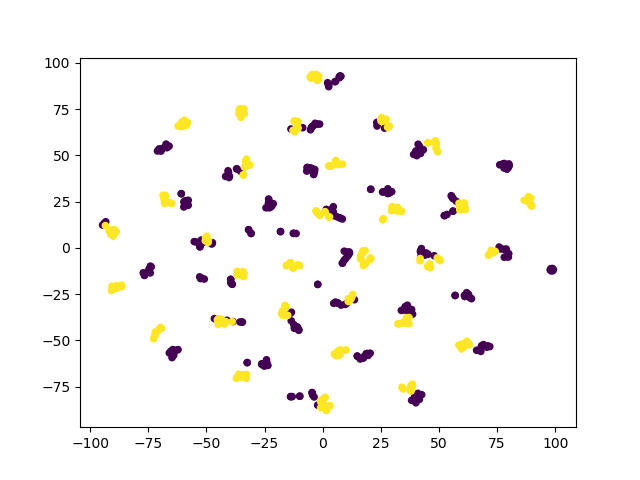}}
		\subfigure[Office-31 Amazon$\rightarrow$Webcam]{
			\label{fig84}
			\includegraphics[width=2.2in]{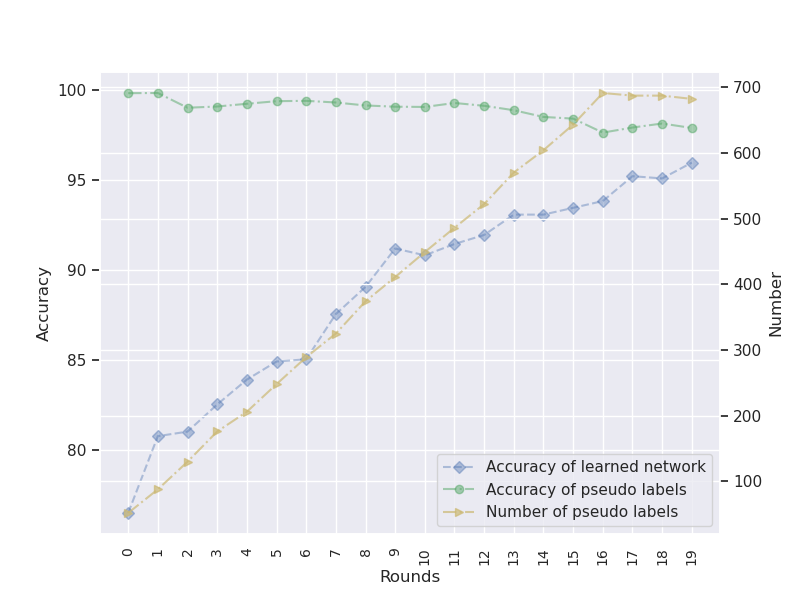}}
		\subfigure[Office-Home Product$\rightarrow$Clipart]{
			\label{fig85}
			\includegraphics[width=2.2in]{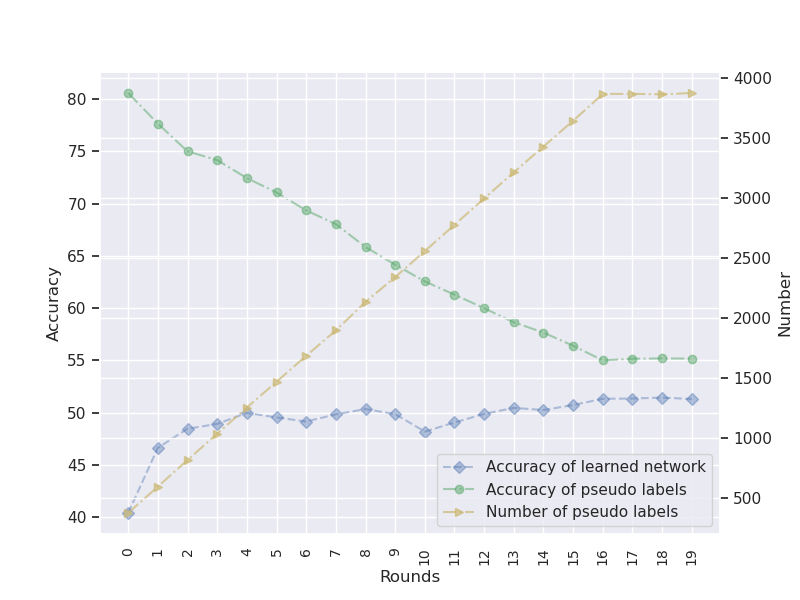}}
		\caption{(a)-(c): T-SNE visualization of features generated by Source-only, CBST, and HCRPL (purple: source, yellow: target). The result is obtained on Office-31 A $\rightarrow$ W under the UDA setting using ResNet-50. (d)-(e): Comparison of the actual accuracy of pseudo labels and learned network accuracy during training. 
		}\label{fig8}
	\end{figure*}

	\begin{table}
		\small
		\centering
		\caption{Results on the Office-31 datasets under the SSDA setting(\%)}	
		\begin{tabular}{llllll}
			\toprule
			\multirow{2}{*}{Network}  & \multirow{2}{*}{Method} & \multicolumn{2}{c}{W $\rightarrow$ A} & \multicolumn{2}{c}{D $\rightarrow$ A} \\
			~ & ~ & 1-shot & 3-shot & 1-shot & 3-shot \\
			\midrule
			\multirow{8}{*}{AlexNet} & S+T & 50.4 & 61.2 & 50.0 & 62.4 \\
			~ & DANN \cite{ganin2016domain} & 57.0 & 64.4 & 54.5 & 65.2 \\
			~ & ADR \cite{saito2017adversarial} & 50.2 & 61.2 & 50.9 & 61.4 \\
			~ & CDAN \cite{long2018conditional} & 50.4 & 60.3 & 48.5 & 61.4 \\
			~ & ENT \cite{grandvalet2005semi} & 50.7 & 64.0 & 50.0 & 66.2 \\
			~ & MME \cite{saito2019semi} & 57.2 & 67.3 & 55.8 & 67.8 \\
			~ & CBST \cite{zou2018domain} & 57.5 & 66.0 & 54.8 & 63.9 \\
			~ & HCRPL & \textbf{63.2} & \textbf{69.9} & \textbf{61.4} & \textbf{70.0} \\
			\midrule
			\midrule
			\multirow{8}{*}{VGG} & S+T & 57.4 & 62.9 & 68.7 & 73.3 \\
			~ & ENT \cite{grandvalet2005semi}  & 51.6 & 64.8 & 70.6 & 75.3 \\
			~ & CDAN \cite{long2018conditional} & 55.8 & 61.8 & 65.9 & 72.9 \\
			~ & ADR \cite{saito2017adversarial} & 57.4 & 63.0 & 69.4 & 73.7 \\
			~ & DANN \cite{ganin2016domain} & 60.0 & 63.9 & 69.8 & 75.0 \\
			~ & MME \cite{saito2019semi} & 62.7 & 67.6 & 73.4 & 77.0 \\
			~ & CBST \cite{zou2018domain} & 71.4 & 76.6 & 70.8 & 76.2 \\
			~ & HCRPL & \textbf{74.6} & \textbf{77.2} & \textbf{74.0} & \textbf{77.8} \\
			\bottomrule
		\end{tabular}
		\label{table11}
	\end{table}
	
	\subsubsection{Semi-supervised domain adaptation} We show results on Office-31 and Office-Home under the SSDA setting. As shown in Table \ref{table11} and \ref{table12}, the proposed HCRPL achieves the state-of-the-art performance on all settings (e.g., different networks, different labeled target sample size) and sub-tasks. Specifically, on Office-31 1-shot, HCRPL outperforms MME by $5.8\%$ using AlexNet. As a reference, MME outperforms S+T by $6.3\%$ under the same setting. Training with AlexNet is more challenging than VGG, but HCRPL improves more on AlexNet, which also proves the effectiveness of HCRPL in challenging scenarios. Similarly, under the SSDA setting, HCRPL has a larger improvement on the tasks with larger domain shifts.
	
	\subsubsection{Comparisons with CBST}\ Tables \ref{table1}, \ref{table2}, \ref{table5}, \ref{table11} and \ref{table12} also provide the results of CBST on different tasks. HCRPL not only outperforms on all tasks but also improves performance by a large margin compared to CBST, which implies that HCRPL promotes overall compared to CBST. Specifically, we discover that performance improvements are more visible on hard tasks (for example, {A $\rightarrow$ D} in Office-31), which illustrates that the hard class problem will further deteriorate the performance of pseudo-labeling in the case of large domain shift, and the proposed HCRPL could alleviate it.

	\begin{table}
		\centering
		\caption{Ablation study under four settings. U and S denote UDA and SSDA, respectively; and R and A denote ResNet50 and AlexNet, respectively. 1 means 1-shot. w/o means without.}
		\begin{tabular}{lrrrr}
			\toprule
			\multirow{2}{*}{Methods}  & Cl$\rightarrow$Ar & A$\rightarrow$W & Rw$\rightarrow$Cl & D$\rightarrow$A\\
			~ & UR & UR & SA1 & SV1 \\
			\midrule
			CBST    & 56.5  & 87.0 &  39.7 & 62.4   \\
			HCRPL w/o APC    & 62.3  & 88.6  & 40.4  & 65.1  \\
			HCRPL w/o SE    & 66.6  & 92.5    & 45.3 &  68.8 \\
			HCRPL w/o TE    & 66.2  & 93.1     & 44.7 & 69.1\\
			HCRPL (full)      & \textbf{67.2}  & \textbf{95.9}  & \textbf{46.0}   & \textbf{70.0} \\
			\bottomrule
		\end{tabular}
		\label{table4}
	\end{table}

	\subsection{Ablation Study}
	We conduct ablation studies under four different settings. The results are shown in Table \ref{table4}. As shown, the APC module plays the most important role in HCRPL. The large performance degradation without APC indicates that calibrating the predictions of target samples is effective to improve transfer performance. Besides, SE and TE could further improve performance by improving the reliability of predictions. Meanwhile, it is found that the accuracy of pseudo labels varies similarly without SE or TE, which illustrates that they perform similar roles. Moreover, the performance of CBST is much lower than the proposed methods, which proves that the hard class problem deteriorates transfer performance dramatically and the proposed schemes are effective.

	\begin{figure*}
		\centering
		\subfigure[CM: Source only]{
			\label{fig61}
			\includegraphics[width=0.67\columnwidth]{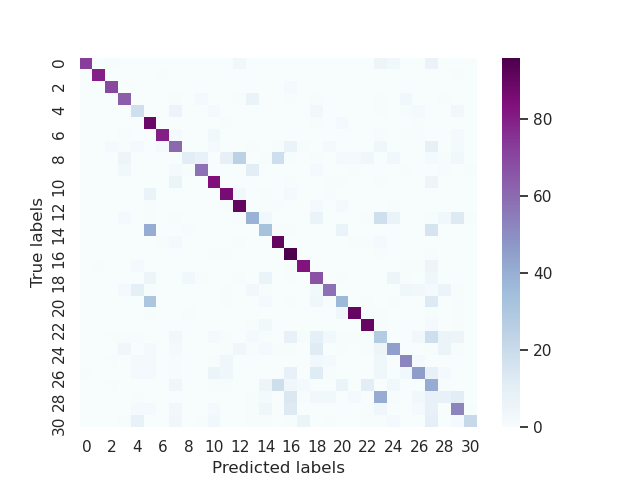}}
		\subfigure[CM: CBST]{
			\label{fig62}
			\includegraphics[width=0.67\columnwidth]{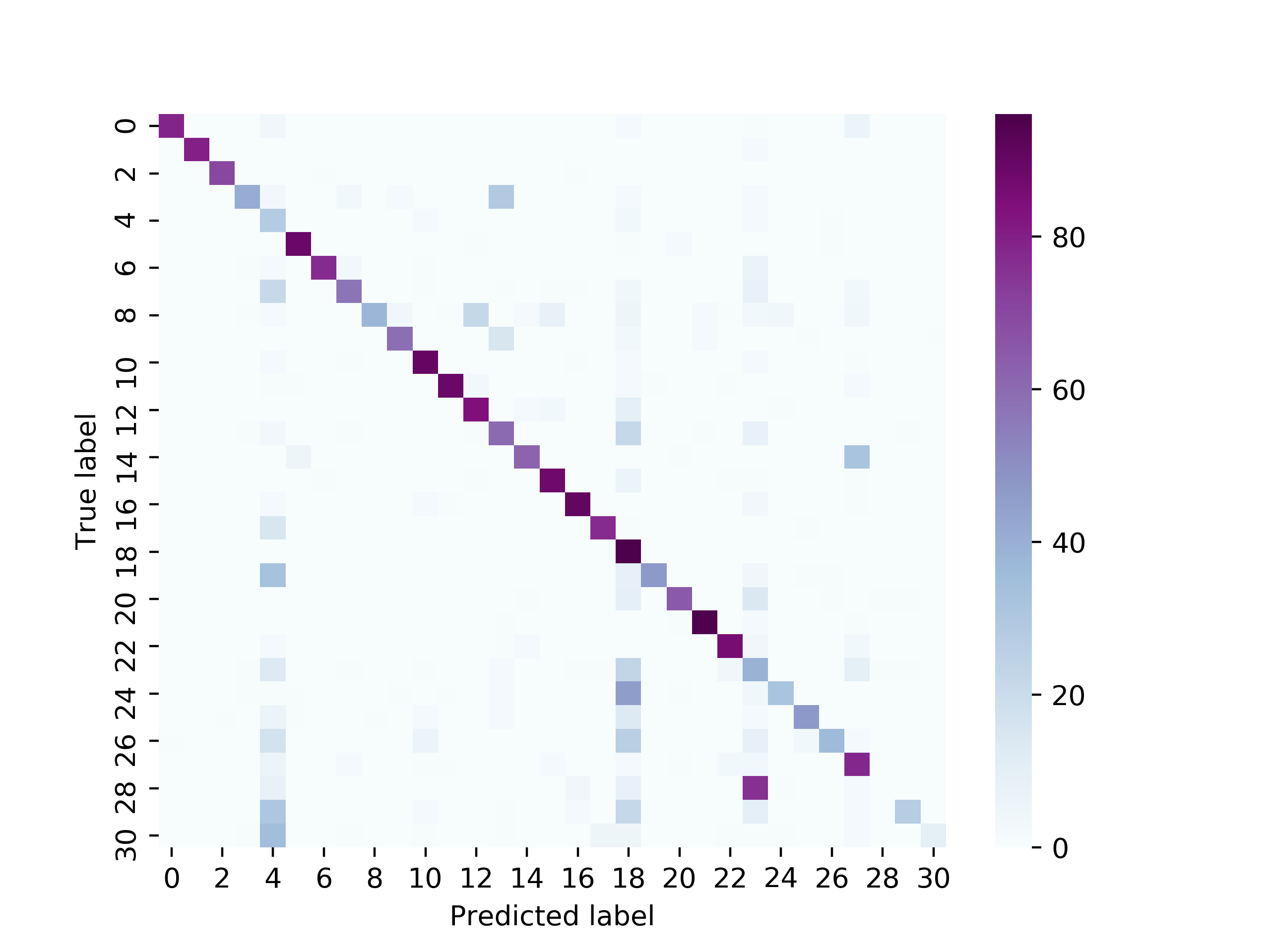}}
		\subfigure[CM: HCRPL]{
			\label{fig63}
			\includegraphics[width=0.67\columnwidth]{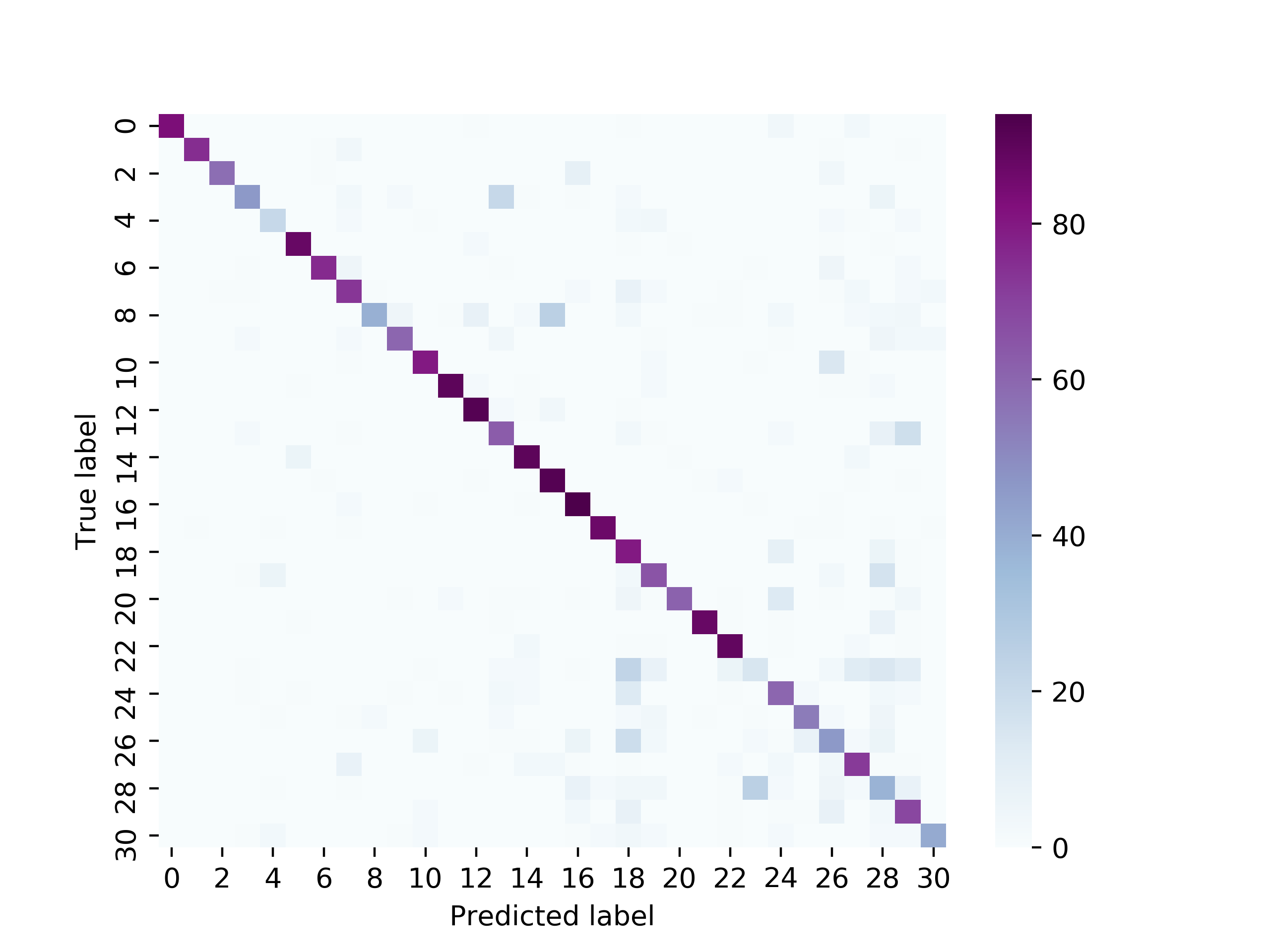}}
		\subfigure[Precision]{
			\label{fig64}
			\includegraphics[width=0.67\columnwidth]{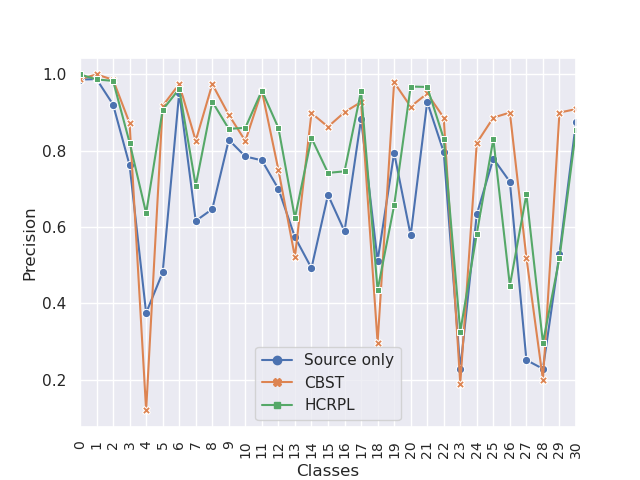}}
		\subfigure[Recall]{
			\label{fig65}
			\includegraphics[width=0.67\columnwidth]{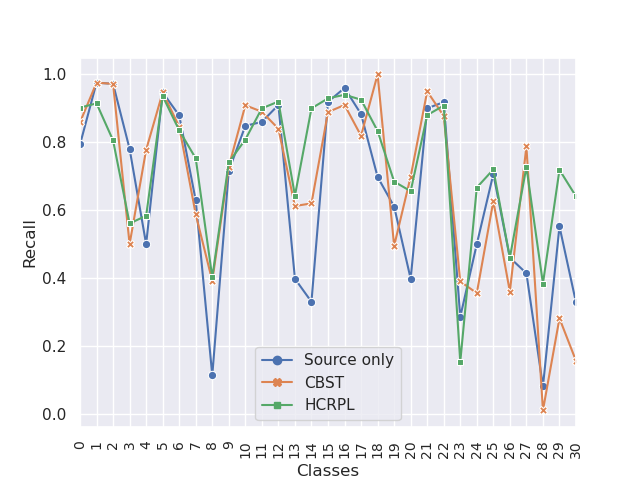}}
		\subfigure[F1-score]{
			\label{fig66}
			\includegraphics[width=0.67\columnwidth]{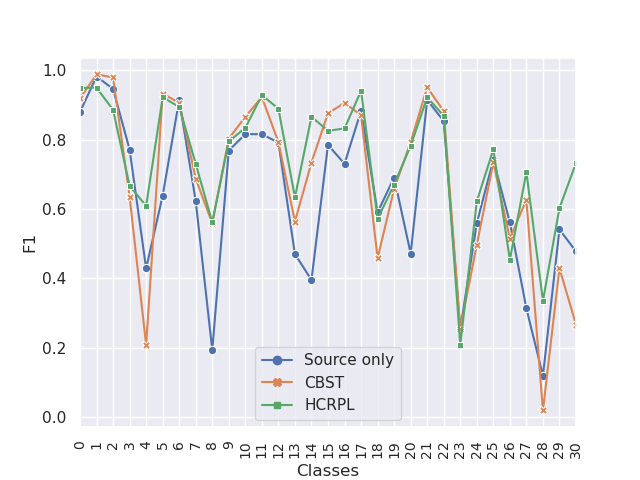}}
		\caption{(a)-(c) Confusion Matrix (CM) visualization for Source only, CBST, and HCRPL. (d)-(f):  Precision, recall, and f1-score evaluated on three different models, Source-only, CBST, and HCRPL. The result is obtained on Office-31 W $\rightarrow$ A under the UDA setting using ResNet-50. To better visualize results, we arrange the categories in alphabetic order.
		}\label{fig6}
	\end{figure*}

	\subsection{Analysis}
	
	\begin{center}
		\begin{table*}[htbp]
			\small
			\centering
			\caption{Comparison of different prior class proportion on Office-31 under the UDA setting (\%). S denotes the marginal class proportion of the source domain. T denotes the marginal class proportion of the target domain.}
			\setlength{\tabcolsep}{-2mm}{
				\begin{tabular*}{\hsize}{@{}@{\extracolsep{\fill}}cccccccc@{}}
					\toprule
					prior class proportion & A $\rightarrow$ W & D $\rightarrow$ W & W $\rightarrow$ D & A $\rightarrow$ D & D $\rightarrow$ A & W $\rightarrow$ A & Avg\\
					\hline
					S & 95.9$\pm$0.2 & 98.7$\pm$0.1 & \textbf{100.0}$\pm$0.0 & 94.3$\pm$0.2 & 75.0$\pm$0.4 & 75.4$\pm$0.4 & 89.9\\
					\hline
					T & \textbf{96.3}$\pm$0.1 & \textbf{98.9}$\pm$0.1 & \textbf{100.0}$\pm$0.0 & \textbf{94.9}$\pm$0.2 & \textbf{75.5}$\pm$0.3 & \textbf{76.2}$\pm$0.3 & \textbf{90.3}\\
					\bottomrule
					\vspace{-0.55cm}
			\end{tabular*}}
			\label{tab1}
		\end{table*}
	\end{center}
	
	{\subsubsection{The effect of different prior class proportions.} \label{pcp} We respectively set marginal class proportion of the source and target domains as prior class proportion and report the results in Table \ref{tab1}. The overall transfer accuracy of the former is only $0.4\%$ lower than the latter, which supports that adopting the former to calibrate predictions is reasonable when the latter is unknown.}
	
	{\subsubsection{Pseudo-labeling Accuracy.} We report the accuracy of pseudo labels and learned network during training on Office-31 A$\rightarrow$W and Office-Home Pr$\rightarrow$Cl under the UDA setting in Figure \ref{fig84} and \ref{fig85}, respectively. We found that (1) As training processes, test accuracy increases steadily, which illuminates the stability of our approach, and hence it can adapt to various scenarios better. (2) Test accuracy maintains a tight relationship with accuracy and the number of pseudo labels. In Office-31 A$\rightarrow$W, the accuracy of pseudo labels keeps stable, and the number increases steadily. Meanwhile, the test accuracy can keep step with the number of pseudo labels. Office-Home Pr$\rightarrow$Cl is a quite challenging task, and our approach also  can extract positive information and improve test accuracy in the process of pseudo-labeling.}

	{\subsubsection{Exploring the hard classes.} We report confusion matrix, precision, recall, and f1-score of source only, CBST, and our approach in Figure \ref{fig6}. We first compare the results of Source only and CBST. Although precision, recall, and f1-score of majority classes are better to a certain extent, the worse performances among all classes may deteriorate further. Here, we focus on the performance of 28-th class in the confusion matrix. The majority of samples belonging to this class are misclassified into the 23-th class in the case of training on the source domain only. After pseudo-labeling, predictions more center on the 23-th class, which results from the misguidance of false predictions with high confidence. The results of precision, recall, and f1-score also confirm this conclusion. Furthermore, we found that the predictive class proportion of easy classes, such as 4-th and 18-th classes, are higher after applying pseudo-labeling (shown in Figures \ref{fig61} and \ref{fig62}), which results in the decline of precision for these classes, which also is the drawback of pseudo-labeling.}
	
	{We further analyze the performance of our HCRPL from the aspect of hard classes. First, the HCRPL avoids predictions degenerate into few classes, which illustrates the HCRPL can control the predictive class proportion of each class in a reasonable interval and improve the precision of the easy classes. Second, we found the precision of majority classes, is better than CBST and source only. The recall is flat or slightly better than CBST and source only. Overall, class-level performance exceeds CBST and source only. Especially for hard classes, such as 28-th, and 30-th classes, the performance of HCRPL is superior apparently, which demonstrates that our approach can indeed alleviate the hard class problem.}

	\subsubsection{Feature Visualization.} We train ResNet-50, CBST, and the proposed HCRPL on Office-31 A $\rightarrow$ W under the UDA setting and plot the learned features with t-SNE \cite{maaten2008visualizing} in Figure \ref{fig8} (a-c), respectively. The purple and yellow points represent the learned features of the source domain and target domains, respectively. As mentioned in section \ref{sec1}, pseudo-labeling is a promising paradigm for aligning and separating the class conditional distributions of various domains. Therefore, CBST and HCRPL could promote learning target-discriminative representations and aligning class conditional distributions. Furthermore, HCRPL could learn more target-discriminative representations due to improving the accuracy of pseudo labels.
	
	\begin{figure}
		\centering
		\includegraphics[width=3.1in]{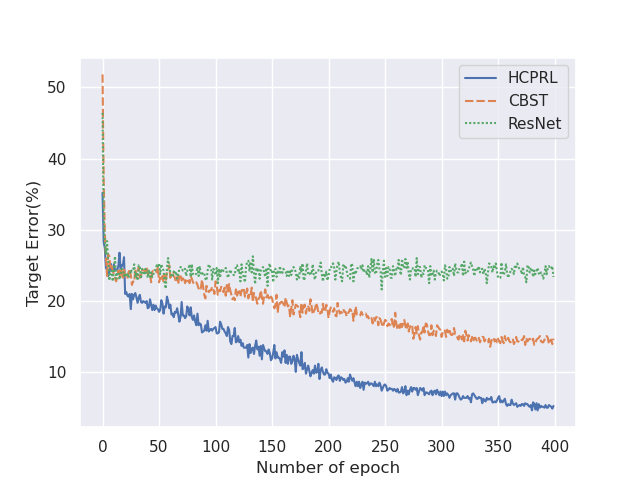}
		\caption{Test accuracy over iterations. The result is obtained on Office-31 A $\rightarrow$ W under the UDA setting using ResNet-50. ResNet means that we train the model without DA and only use the source domain as training data. 
		}\label{fig9}
	\end{figure}
	
	\subsubsection{Convergence.} The convergence of ResNet, CBST, and the proposed HCRPL can be demonstrated by the error rates in the target domain on Office-31 A $\rightarrow$ W under the UDA setting. As shown in Figure \ref{fig9}, the following observations can be made: 1) CBST and HCRPL are more stable than ResNet (baseline) due to alleviating overfitting to the source domain. 2) The target error decreases gradually learned by CBST and HCRPL because the pseudo-labeling methods progressively generate more pseudo labels and learn more of the target-discriminative representations as the training progresses. 3) The disparity of the target error between CBST and HCRPL increases gradually, which indicates that APC, SE, and TE effectively improve the accuracy of pseudo labels.

	\begin{table}[htbp]
		\centering
		\setlength{\belowcaptionskip}{-0.1cm}
		\setlength{\abovecaptionskip}{3pt}
		\caption{Comparison of different EMA momentum $\alpha$ and sharpening temperature $T$ on A$\rightarrow$W.}
		\begin{tabular}{|c|c|c|c|c|c|c|}
			\hline
			$\alpha$ & 0.0 & 0.8 & 0.9 & 0.95 & 0.97 & 0.99 \\
			\hline
			Acc ($\%$) & 92.5 & 93.1 & 95.2 & \textbf{95.9} & 94.5 & 93.1 \\
			\hline
			\hline
			$T$ & 0.2 & 0.3 & 0.5 & 0.7 & 0.8 & 1.0 \\
			\hline
			Acc ($\%$) & 94.7 & 95.7 & \textbf{95.9} & 93.8 & 93.8 & 92.3 \\
			\hline
		\end{tabular}
		\label{tab2}
	\end{table}

	{\subsubsection{Hyperparameter Sensitivity}
		We then investigate the sensitivity of HCRPL for different choices of hyperparameters: EMA momentum $\alpha$ and sharpening temperature $T$. The results are reported in Table \ref{tab2}. For EMA momentum $\alpha$, transfer accuracy fluctuates slightly when $\alpha \in [0.9, 0.97]$. However, when the value of $\alpha$ does not belong to this range, transfer accuracy degrades a lot. For sharpening temperature $T$, the smaller $T$ achieves better performance than the larger one, which means the predictions with lower entropy are more accurate. Meanwhile, the minimal value of $T$ also degrades performance.}

	\begin{table*}
		\centering
		\caption{Results on Office-Home dataset under the SSDA setting(\%)}	
		\begin{tabular}{lllllllllllllll}
			\toprule
			\multirow{2}{*}{Network} & \multirow{2}{*}{Method} & Ar & Ar & Ar & Cl & Cl & Cl & Pr & Pr & Pr & Rw & Rw & Rw & \multirow{2}{*}{Avg}\\
			~ & ~ & Cl & Pr & Rw & Ar & Pr & Rw & Ar & Cl & Rw & Ar & Cl & Pr & ~\\
			\midrule
			\multicolumn{15}{c}{\textbf{One-shot}} \\
			\midrule
			\multirow{8}{*}{AlexNet} & S+T & 37.5 & 63.1 & 44.8 & 54.3 & 31.7 & 31.5 & 48.8 & 31.1 & 53.3 & 48.5 & 33.9 & 50.8 & 44.1 \\
			~ & DANN \cite{ganin2016domain} & 42.5 & 64.2 & 45.1 & 56.4 & 36.6 & 32.7 & 43.5 & 34.4 & 51.9 & 51.0 & 33.8 & 49.4 & 45.1 \\
			~ & ADR \cite{saito2017adversarial} & 37.8 & 63.5 & 45.4 & 53.5 & 32.5 & 32.2 & 49.5 & 31.8 & 53.4 & 49.7 & 34.2 & 50.4 & 44.5 \\
			~ & CDAN \cite{long2018conditional} & 36.1 & 62.3 & 42.2 & 52.7 & 28.0 & 27.8 & 48.7 & 28.0 & 51.3 & 41.0 & 26.8 & 49.9 & 41.2 \\
			~ & ENT \cite{grandvalet2005semi} & 26.8 & 65.8 & 45.8 & 56.3 & 23.5 & 21.9 & 47.4 & 22.1 & 53.4 & 30.8 & 18.1 & 53.6 & 38.8 \\
			~ & MME \cite{saito2019semi} & 42.0 & 69.6 & 48.3 & 58.7 & 37.8 & 34.9 & 52.5 & 36.4 & 57.0 & 54.1 & 39.5 & 59.1 & 49.2 \\
			~ & CBST \cite{zou2018domain} & 39.7 & 69.1 & 46.0 & 59.3 & 31.6 & 33.7 & 54.6 & 32.5 & 57.3 & 53.2 & 36.0 & 57.4 & 47.5 \\
			~ & HCRPL & \textbf{46.0} & \textbf{73.3} & \textbf{48.8} & \textbf{63.8} & \textbf{38.7} & \textbf{38.4} & \textbf{58.1} & \textbf{37.4} & \textbf{59.7} & \textbf{61.0} & \textbf{40.1} & \textbf{62.2} & \textbf{52.3} \\
			\midrule
			\midrule
			\multirow{8}{*}{VGG} & S+T & 39.5 & 75.3 & 61.2 & 71.6 & 37.0 & 52.0 & 63.6 & 37.5 & 69.5 & 64.5 & 51.4 & 65.9 & 57.4 \\
			~ & DANN \cite{ganin2016domain} & 52.0 & 75.7 & 62.7 & 72.7 & 45.9 & 51.3 & 64.3 & 44.4 & 68.9 & 64.2 & 52.3 & 65.3 & 60.0 \\
			~ & ADR \cite{saito2017adversarial} & 39.7 & 76.2 & 60.2 & 71.8 & 37.2 & 51.4 & 63.9 & 39.0 & 68.7 & 64.8 & 50.0 & 65.2 & 57.4 \\
			~ & CDAN \cite{long2018conditional} & 43.3 & 75.7 & 60.9 & 69.6 & 37.4 & 44.5 & 67.7 & 39.8 & 64.8 & 58.7 & 41.6 & 66.2 & 55.8 \\
			~ & ENT \cite{grandvalet2005semi} & 23.7 & 77.5 & 64.0 & 74.6 & 21.3 & 44.6 & 66.0 & 22.4 & 70.6 & 62.1 & 25.1 & 67.7 & 51.6 \\
			~ & MME \cite{saito2019semi} & 49.1 & 78.7 & 65.1 & 74.4 & 46.2 & 56.0 & 68.6 & 45.8 & 72.2 & 68.0 & 57.5 & 71.3 & 62.7 \\
			~ & CBST \cite{zou2018domain} & 42.2 & 78.9 & 62.2 & 75.0 & 39.5 & 52.8 & 70.6 & 40.4 & 73.2 & 68.8 & 54.1 & 70.7 & 60.7 \\
			~ & HCRPL & \textbf{53.1} & \textbf{82.0} & \textbf{66.3} & \textbf{77.1} & \textbf{49.0} & \textbf{57.8} & \textbf{76.3} & \textbf{47.4} & \textbf{75.6} & \textbf{73.5} & \textbf{58.3} & \textbf{73.8} & \textbf{65.9} \\
			\midrule
			\multicolumn{15}{c}{\textbf{Three-shot}} \\
			\midrule
			\multirow{8}{*}{AlexNet} & S+T & 44.6 & 66.7 & 47.7 & 57.8 & 44.4 & 36.1 & 57.6 & 38.8 & 57.0 & 54.3 & 37.5 & 57.9 & 50.0 \\
			~ & DANN \cite{ganin2016domain} & 47.2 & 66.7 & 46.6 & 58.1 & 44.4 & 36.1 & 57.2 & 39.8 & 56.6 & 54.3 & 38.6 & 57.9 & 50.3 \\
			~ & ADR \cite{saito2017adversarial} & 45.0 & 66.2 & 46.9 & 57.3 & 38.9 & 36.3 & 57.5 & 40.0 & 57.8 & 53.4 & 37.3 & 57.7 & 49.5 \\
			~ & CDAN \cite{long2018conditional} & 41.8 & 69.9 & 43.2 & 53.6 & 35.8 & 32.0 & 56.3 & 34.5 & 53.5 & 49.3 & 27.9 & 56.2 & 46.2 \\
			~ & ENT \cite{grandvalet2005semi} & 44.9 & 70.4 & 47.1 & 60.3 & 41.2 & 34.6 & 60.7 & 37.8 & 60.5 & 58.0 & 31.8 & 63.4 & 50.9 \\
			~ & MME \cite{saito2019semi} & 51.2 & 73.0 & 50.3 & 61.6 & 47.2 & 40.7 & 63.9 & 43.8 & 61.4 & 59.9 & 44.7 & 64.7 & 55.2 \\
			~ & CBST \cite{zou2018domain} & 45.3 & 72.6 & 48.4 & 62.3 & 40.3 & 37.3 & 64.2 & 40.7 & 61.6 & 58.6 & 39.9 & 62.6 & 52.8 \\
			~ & HCRPL & \textbf{53.5} & \textbf{75.0} & \textbf{51.4} & \textbf{64.9} & \textbf{46.0} & \textbf{42.4} & \textbf{65.4} & \textbf{45.8} & \textbf{63.4} & \textbf{63.1} & \textbf{41.9} & \textbf{67.6} & \textbf{56.7} \\
			\midrule
			\midrule
			\multirow{8}{*}{VGG} & S+T & 49.6 & 78.6 & 63.6 & 72.7 & 47.2 & 55.9 & 69.4 & 47.5 & 73.4 & 69.7 & 56.2 & 70.4 & 62.9 \\
			~ & DANN \cite{ganin2016domain} & 56.1 & 77.9 & 63.7 & 73.6 & 52.4 & 56.3 & 69.5 & 50.0 & 72.3 & 68.7 & 56.4 & 69.8 & 63.9 \\
			~ & ADR \cite{saito2017adversarial} & 49.0 &  78.1 & 62.8 & 73.6 & 47.8 & 55.8 & 69.9 & 49.3 & 73.3 & 69.3 & 56.3 & 71.4 & 63.0 \\
			~ & CDAN \cite{long2018conditional} & 50.2 & 80.9 & 62.1 & 70.8 & 45.1 & 50.3 & 74.7 & 46.0 & 71.4 & 65.9 & 52.9 & 71.2 & 61.8 \\
			~ & ENT \cite{grandvalet2005semi} & 48.3 & 81.6 & 65.5 & 76.6 & 46.8 & 56.9 & 73.0 & 44.8 & 75.3 & 72.9 & 59.1 & 77.0 & 64.8 \\
			~ & MME \cite{saito2019semi} & 56.9 & 82.9 & 65.7 & 76.7 & 53.6 & 59.2 & 75.7 & 54.9 & 75.3 & 72.9 & 61.1 & 76.3 & 67.6 \\
			~ & MME \cite{saito2019semi} & 56.9 & 82.9 & 65.7 & 76.7 & 53.6 & 59.2 & 75.7 & 54.9 & 75.3 & 72.9 & 61.1 & 76.3 & 67.6 \\
			~ & CBST \cite{zou2018domain} & 52.2 & 81.6 & 64.8 & 75.5 & 48.9 & 55.7 & 75.1 & 51.4 & 76.3 & 72.0 & 57.5 & 74.7 & 65.5 \\
			~ & HCRPL & \textbf{59.7} & \textbf{84.7} & \textbf{68.7} & \textbf{78.5} & \textbf{55.3} & \textbf{61.7} & \textbf{77.6} & \textbf{55.5} & \textbf{78.6} & \textbf{76.1} & \textbf{62.2} & \textbf{79.0} & \textbf{69.8} \\
			\bottomrule
		\end{tabular}
		\label{table12}
	\end{table*}

	\section{Discussion and Conclusion}
	
	{\subsection{Strength}}
	{The major contribution of this paper is unraveling the hard class problem, which is always ignored in the existing pseudo-labeling methods yet is crucial in some practical scenarios. Compared with the existing pseudo-labeling methods, our approach improves not only overall performance but only the worst performance among all classes.}
	
	{\subsection{Weakness}}
	{The introduction of prior knowledge alleviates the hard class problem effectively. However, it also limits the application of our approach to a certain extent. Our basic assumption is that the source and target domains should have a similar label proportion, which is invalid in some applications, such as partial domain adaptation \cite{cao2018partial}, open set domain adaptation \cite{panareda2017open}. In such variants of DA, applying APC obviously will miscalibrate predictions and result in massive false pseudo labels. Therefore, accurately inferring the marginal class distribution of the target domain should be further studied in future jobs. On the other hand, the proposed HCRPL can effectively improve the precision of hard classes but has little impact on the recall of them, which also is important in practical applications.}

	\subsection{Conclusion}
	Pseudo-labeling is a promising paradigm for solving the DA problem. In this paper, we first revealed the hard class problem may occur in DA and is harmful to the applications of DA. To alleviate the hard class problem, we proposed APC to calibrate predictions according to the difficulty degree for each class. Furthermore, we introduced SE and TE to improve model robustness for target samples, especially for ones belonging to hard classes. In experiments, we demonstrated that HCRPL achieved good results and outperformed some state-of-the-art methods with considerable margins under the UDA setting. Meanwhile, HCRPL is suitable for the SSDA setting because it can learn target-discriminative representations. Experimental analysis shows that the proposed schemes can indeed alleviate the hard class problem and improve the accuracy of pseudo labels.

	\section{Acknowledgments}
	
	The work is supported in part by National Natural Science Foundation of China under Grants 81671766, 619713-69, U19B2031 U1605252, 61671309, in part by Open Fund of Science and Technology on Automatic Target Recognition Laboratory 6142503190202, in part by Fundamental Research Funds for the Central Universities 20720180059, 20-720190116, 20720200003, in part by Tencent Open Fund.

	\bibliographystyle{cas-model2-names}

	\bibliography{KBS}

\begin{thebibliography}{76}
\expandafter\ifx\csname natexlab\endcsname\relax\def\natexlab#1{#1}\fi
\providecommand{\url}[1]{\texttt{#1}}
\providecommand{\href}[2]{#2}
\providecommand{\path}[1]{#1}
\providecommand{\DOIprefix}{doi:}
\providecommand{\ArXivprefix}{arXiv:}
\providecommand{\URLprefix}{URL: }
\providecommand{\Pubmedprefix}{pmid:}
\providecommand{\doi}[1]{\href{http://dx.doi.org/#1}{\path{#1}}}
\providecommand{\Pubmed}[1]{\href{pmid:#1}{\path{#1}}}
\providecommand{\bibinfo}[2]{#2}
\ifx\xfnm\relax \def\xfnm[#1]{\unskip,\space#1}\fi
\bibitem[{Amodei et~al.(2015)Amodei, Ananthanarayanan, Anubhai, Bai and
  Zhu}]{Amodei2015Deep}
\bibinfo{author}{Amodei, D.}, \bibinfo{author}{Ananthanarayanan, S.},
  \bibinfo{author}{Anubhai, R.}, \bibinfo{author}{Bai, J.},
  \bibinfo{author}{Zhu, Z.}, \bibinfo{year}{2015}.
\newblock \bibinfo{title}{Deep speech 2: End-to-end speech recognition in
  english and mandarin}.
\newblock \bibinfo{journal}{Computer Science} .
\bibitem[{Ben{-}David et~al.(2010)Ben{-}David, Blitzer, Crammer, Kulesza,
  Pereira and Vaughan}]{DBLP:journals/ml/Ben-DavidBCKPV10}
\bibinfo{author}{Ben{-}David, S.}, \bibinfo{author}{Blitzer, J.},
  \bibinfo{author}{Crammer, K.}, \bibinfo{author}{Kulesza, A.},
  \bibinfo{author}{Pereira, F.}, \bibinfo{author}{Vaughan, J.W.},
  \bibinfo{year}{2010}.
\newblock \bibinfo{title}{A theory of learning from different domains}.
\newblock \bibinfo{journal}{Machine Learning} \bibinfo{volume}{79},
  \bibinfo{pages}{151--175}.
\bibitem[{Berthelot et~al.(2019a)Berthelot, Carlini, Cubuk, Kurakin, Sohn,
  Zhang and Raffel}]{berthelot2019remixmatch}
\bibinfo{author}{Berthelot, D.}, \bibinfo{author}{Carlini, N.},
  \bibinfo{author}{Cubuk, E.D.}, \bibinfo{author}{Kurakin, A.},
  \bibinfo{author}{Sohn, K.}, \bibinfo{author}{Zhang, H.},
  \bibinfo{author}{Raffel, C.}, \bibinfo{year}{2019}a.
\newblock \bibinfo{title}{Remixmatch: Semi-supervised learning with
  distribution alignment and augmentation anchoring}.
\newblock \bibinfo{journal}{arXiv preprint arXiv:1911.09785} .
\bibitem[{Berthelot et~al.(2019b)Berthelot, Carlini, Goodfellow, Papernot,
  Oliver and Raffel}]{berthelot2019mixmatch}
\bibinfo{author}{Berthelot, D.}, \bibinfo{author}{Carlini, N.},
  \bibinfo{author}{Goodfellow, I.}, \bibinfo{author}{Papernot, N.},
  \bibinfo{author}{Oliver, A.}, \bibinfo{author}{Raffel, C.A.},
  \bibinfo{year}{2019}b.
\newblock \bibinfo{title}{Mixmatch: A holistic approach to semi-supervised
  learning}, in: \bibinfo{booktitle}{Advances in Neural Information Processing
  Systems}, pp. \bibinfo{pages}{5050--5060}.
\bibitem[{Cao et~al.(2018)Cao, Ma, Long and Wang}]{cao2018partial}
\bibinfo{author}{Cao, Z.}, \bibinfo{author}{Ma, L.}, \bibinfo{author}{Long,
  M.}, \bibinfo{author}{Wang, J.}, \bibinfo{year}{2018}.
\newblock \bibinfo{title}{Partial adversarial domain adaptation}, in:
  \bibinfo{booktitle}{Proceedings of the European Conference on Computer Vision
  (ECCV)}, pp. \bibinfo{pages}{135--150}.
\bibitem[{Chen et~al.(2019a)Chen, Xie, Huang, Rong, Ding, Huang, Xu and
  Huang}]{chen2019progressive}
\bibinfo{author}{Chen, C.}, \bibinfo{author}{Xie, W.}, \bibinfo{author}{Huang,
  W.}, \bibinfo{author}{Rong, Y.}, \bibinfo{author}{Ding, X.},
  \bibinfo{author}{Huang, Y.}, \bibinfo{author}{Xu, T.},
  \bibinfo{author}{Huang, J.}, \bibinfo{year}{2019}a.
\newblock \bibinfo{title}{Progressive feature alignment for unsupervised domain
  adaptation}, in: \bibinfo{booktitle}{Proceedings of the IEEE Conference on
  Computer Vision and Pattern Recognition}, pp. \bibinfo{pages}{627--636}.
\bibitem[{Chen et~al.(2020a)Chen, Zheng, Ding, Huang and
  Dou}]{chen2020harmonizing}
\bibinfo{author}{Chen, C.}, \bibinfo{author}{Zheng, Z.}, \bibinfo{author}{Ding,
  X.}, \bibinfo{author}{Huang, Y.}, \bibinfo{author}{Dou, Q.},
  \bibinfo{year}{2020}a.
\newblock \bibinfo{title}{Harmonizing transferability and discriminability for
  adapting object detectors}, in: \bibinfo{booktitle}{Proceedings of the
  IEEE/CVF Conference on Computer Vision and Pattern Recognition}, pp.
  \bibinfo{pages}{8869--8878}.
\bibitem[{Chen et~al.(2021)Chen, Zheng, Huang, Ding and Yu}]{chen2021I3NET}
\bibinfo{author}{Chen, C.}, \bibinfo{author}{Zheng, Z.},
  \bibinfo{author}{Huang, Y.}, \bibinfo{author}{Ding, X.}, \bibinfo{author}{Yu,
  Y.}, \bibinfo{year}{2021}.
\newblock \bibinfo{title}{I3net: Implicit instance-invariant network for
  adapting one-stage object detectors}, in: \bibinfo{booktitle}{IEEE Conference
  on Computer Vision and Pattern Recognition (CVPR)}.
\bibitem[{Chen et~al.(2019b)Chen, Wang, Yi, Chen and Zhou}]{chen2019joint}
\bibinfo{author}{Chen, D.D.}, \bibinfo{author}{Wang, Y.}, \bibinfo{author}{Yi,
  J.}, \bibinfo{author}{Chen, Z.}, \bibinfo{author}{Zhou, Z.H.},
  \bibinfo{year}{2019}b.
\newblock \bibinfo{title}{Joint semantic domain alignment and target classifier
  learning for unsupervised domain adaptation}.
\newblock \bibinfo{journal}{arXiv preprint arXiv:1906.04053} .
\bibitem[{Chen et~al.(2020b)Chen, Harandi, Jin and Yang}]{Chen2020DomainAB}
\bibinfo{author}{Chen, S.}, \bibinfo{author}{Harandi, M.},
  \bibinfo{author}{Jin, X.}, \bibinfo{author}{Yang, X.}, \bibinfo{year}{2020}b.
\newblock \bibinfo{title}{Domain adaptation by joint distribution invariant
  projections}.
\newblock \bibinfo{journal}{IEEE Transactions on Image Processing}
  \bibinfo{volume}{29}, \bibinfo{pages}{8264--8277}.
\bibitem[{Deng et~al.(2018)Deng, Zheng and
  Jiao}]{DBLP:journals/corr/abs-1812-00893}
\bibinfo{author}{Deng, W.}, \bibinfo{author}{Zheng, L.}, \bibinfo{author}{Jiao,
  J.}, \bibinfo{year}{2018}.
\newblock \bibinfo{title}{Domain alignment with triplets}.
\newblock \bibinfo{journal}{CoRR} \bibinfo{volume}{abs/1812.00893}.
\bibitem[{Donahue et~al.(2014)Donahue, Jia, Vinyals, Hoffman, Zhang, Tzeng and
  Darrell}]{Donahue2014DeCAFAD}
\bibinfo{author}{Donahue, J.}, \bibinfo{author}{Jia, Y.},
  \bibinfo{author}{Vinyals, O.}, \bibinfo{author}{Hoffman, J.},
  \bibinfo{author}{Zhang, N.}, \bibinfo{author}{Tzeng, E.},
  \bibinfo{author}{Darrell, T.}, \bibinfo{year}{2014}.
\newblock \bibinfo{title}{Decaf: A deep convolutional activation feature for
  generic visual recognition}, in: \bibinfo{booktitle}{ICML}.
\bibitem[{{Fernando} et~al.(2013){Fernando}, {Habrard}, {Sebban} and
  {Tuytelaars}}]{6751479}
\bibinfo{author}{{Fernando}, B.}, \bibinfo{author}{{Habrard}, A.},
  \bibinfo{author}{{Sebban}, M.}, \bibinfo{author}{{Tuytelaars}, T.},
  \bibinfo{year}{2013}.
\newblock \bibinfo{title}{Unsupervised visual domain adaptation using subspace
  alignment}, in: \bibinfo{booktitle}{2013 IEEE International Conference on
  Computer Vision}, pp. \bibinfo{pages}{2960--2967}.
\newblock \DOIprefix\doi{10.1109/ICCV.2013.368}.
\bibitem[{French et~al.(2018)French, Mackiewicz and
  Fisher}]{DBLP:conf/iclr/FrenchMF18}
\bibinfo{author}{French, G.}, \bibinfo{author}{Mackiewicz, M.},
  \bibinfo{author}{Fisher, M.H.}, \bibinfo{year}{2018}.
\newblock \bibinfo{title}{Self-ensembling for visual domain adaptation}, in:
  \bibinfo{booktitle}{6th International Conference on Learning
  Representations}, \bibinfo{publisher}{OpenReview.net}.
\bibitem[{Ganin and Lempitsky(2015)}]{ganin2014unsupervised}
\bibinfo{author}{Ganin, Y.}, \bibinfo{author}{Lempitsky, V.S.},
  \bibinfo{year}{2015}.
\newblock \bibinfo{title}{Unsupervised domain adaptation by backpropagation},
  in: \bibinfo{editor}{Bach, F.R.}, \bibinfo{editor}{Blei, D.M.} (Eds.),
  \bibinfo{booktitle}{Proceedings of the 32nd International Conference on
  Machine Learning}, \bibinfo{publisher}{JMLR.org}. pp.
  \bibinfo{pages}{1180--1189}.
\bibitem[{Ganin et~al.(2016)Ganin, Ustinova, Ajakan, Germain, Larochelle,
  Laviolette, Marchand and Lempitsky}]{ganin2016domain}
\bibinfo{author}{Ganin, Y.}, \bibinfo{author}{Ustinova, E.},
  \bibinfo{author}{Ajakan, H.}, \bibinfo{author}{Germain, P.},
  \bibinfo{author}{Larochelle, H.}, \bibinfo{author}{Laviolette, F.},
  \bibinfo{author}{Marchand, M.}, \bibinfo{author}{Lempitsky, V.S.},
  \bibinfo{year}{2016}.
\newblock \bibinfo{title}{Domain-adversarial training of neural networks}.
\newblock \bibinfo{journal}{J. Mach. Learn. Res.} \bibinfo{volume}{17},
  \bibinfo{pages}{59:1--59:35}.
\bibitem[{Ghifary et~al.(2017)Ghifary, Balduzzi, Kleijn and
  Zhang}]{Ghifary2017ScatterCA}
\bibinfo{author}{Ghifary, M.}, \bibinfo{author}{Balduzzi, D.},
  \bibinfo{author}{Kleijn, W.}, \bibinfo{author}{Zhang, M.},
  \bibinfo{year}{2017}.
\newblock \bibinfo{title}{Scatter component analysis: A unified framework for
  domain adaptation and domain generalization}.
\newblock \bibinfo{journal}{IEEE Transactions on Pattern Analysis and Machine
  Intelligence} \bibinfo{volume}{39}, \bibinfo{pages}{1414--1430}.
\bibitem[{Gong et~al.(2012)Gong, Shi, Sha and Grauman}]{gong2012geodesic}
\bibinfo{author}{Gong, B.}, \bibinfo{author}{Shi, Y.}, \bibinfo{author}{Sha,
  F.}, \bibinfo{author}{Grauman, K.}, \bibinfo{year}{2012}.
\newblock \bibinfo{title}{Geodesic flow kernel for unsupervised domain
  adaptation}, in: \bibinfo{booktitle}{2012 IEEE conference on computer vision
  and pattern recognition}, \bibinfo{organization}{IEEE}. pp.
  \bibinfo{pages}{2066--2073}.
\bibitem[{Goodfellow et~al.(2014)Goodfellow, Pouget-Abadie, Mirza, Xu,
  Warde-Farley, Ozair, Courville and Bengio}]{goodfellow2014generative}
\bibinfo{author}{Goodfellow, I.}, \bibinfo{author}{Pouget-Abadie, J.},
  \bibinfo{author}{Mirza, M.}, \bibinfo{author}{Xu, B.},
  \bibinfo{author}{Warde-Farley, D.}, \bibinfo{author}{Ozair, S.},
  \bibinfo{author}{Courville, A.}, \bibinfo{author}{Bengio, Y.},
  \bibinfo{year}{2014}.
\newblock \bibinfo{title}{Generative adversarial nets}, in:
  \bibinfo{booktitle}{Advances in neural information processing systems}, pp.
  \bibinfo{pages}{2672--2680}.
\bibitem[{Grandvalet and Bengio(2005)}]{grandvalet2005semi}
\bibinfo{author}{Grandvalet, Y.}, \bibinfo{author}{Bengio, Y.},
  \bibinfo{year}{2005}.
\newblock \bibinfo{title}{Semi-supervised learning by entropy minimization},
  in: \bibinfo{booktitle}{Advances in neural information processing systems},
  pp. \bibinfo{pages}{529--536}.
\bibitem[{Gretton et~al.(2012)Gretton, Borgwardt, Rasch, Sch{\"{o}}lkopf and
  Smola}]{gretton2012kernel}
\bibinfo{author}{Gretton, A.}, \bibinfo{author}{Borgwardt, K.M.},
  \bibinfo{author}{Rasch, M.J.}, \bibinfo{author}{Sch{\"{o}}lkopf, B.},
  \bibinfo{author}{Smola, A.J.}, \bibinfo{year}{2012}.
\newblock \bibinfo{title}{A kernel two-sample test}.
\newblock \bibinfo{journal}{J. Mach. Learn. Res.} \bibinfo{volume}{13},
  \bibinfo{pages}{723--773}.
\bibitem[{He et~al.(2016)He, Zhang, Ren and Sun}]{he2016deep}
\bibinfo{author}{He, K.}, \bibinfo{author}{Zhang, X.}, \bibinfo{author}{Ren,
  S.}, \bibinfo{author}{Sun, J.}, \bibinfo{year}{2016}.
\newblock \bibinfo{title}{Deep residual learning for image recognition}, in:
  \bibinfo{booktitle}{Proceedings of the IEEE conference on computer vision and
  pattern recognition}, pp. \bibinfo{pages}{770--778}.
\bibitem[{Hu et~al.(2020)Hu, Kan, Shan and Chen}]{hu2020unsupervised}
\bibinfo{author}{Hu, L.}, \bibinfo{author}{Kan, M.}, \bibinfo{author}{Shan,
  S.}, \bibinfo{author}{Chen, X.}, \bibinfo{year}{2020}.
\newblock \bibinfo{title}{Unsupervised domain adaptation with hierarchical
  gradient synchronization}, in: \bibinfo{booktitle}{Proceedings of the
  IEEE/CVF Conference on Computer Vision and Pattern Recognition}, pp.
  \bibinfo{pages}{4043--4052}.
\bibitem[{Jing-jing et~al.(2019)Jing-jing, Mengmeng, Ke, Lei and
  Tao}]{Jingjing2019LocalityPJ}
\bibinfo{author}{Jing-jing, L.}, \bibinfo{author}{Mengmeng, J.},
  \bibinfo{author}{Ke, L.}, \bibinfo{author}{Lei, Z.}, \bibinfo{author}{Tao,
  S.}, \bibinfo{year}{2019}.
\newblock \bibinfo{title}{Locality preserving joint transfer for domain
  adaptation}.
\newblock \bibinfo{journal}{arXiv: Computer Vision and Pattern Recognition} .
\bibitem[{Kang et~al.(2019)Kang, Jiang, Yang and
  Hauptmann}]{kang2019contrastive}
\bibinfo{author}{Kang, G.}, \bibinfo{author}{Jiang, L.}, \bibinfo{author}{Yang,
  Y.}, \bibinfo{author}{Hauptmann, A.G.}, \bibinfo{year}{2019}.
\newblock \bibinfo{title}{Contrastive adaptation network for unsupervised
  domain adaptation}, in: \bibinfo{booktitle}{Proceedings of the IEEE
  Conference on Computer Vision and Pattern Recognition}, pp.
  \bibinfo{pages}{4893--4902}.
\bibitem[{Kim and Kim(2020)}]{Kim2020AttractPA}
\bibinfo{author}{Kim, T.K.}, \bibinfo{author}{Kim, C.}, \bibinfo{year}{2020}.
\newblock \bibinfo{title}{Attract, perturb, and explore: Learning a feature
  alignment network for semi-supervised domain adaptation}, in:
  \bibinfo{booktitle}{ECCV}.
\bibitem[{Krizhevsky et~al.(2012)Krizhevsky, Sutskever and
  Hinton}]{krizhevsky2012imagenet}
\bibinfo{author}{Krizhevsky, A.}, \bibinfo{author}{Sutskever, I.},
  \bibinfo{author}{Hinton, G.E.}, \bibinfo{year}{2012}.
\newblock \bibinfo{title}{Imagenet classification with deep convolutional
  neural networks}, in: \bibinfo{editor}{Bartlett, P.L.},
  \bibinfo{editor}{Pereira, F.C.N.}, \bibinfo{editor}{Burges, C.J.C.},
  \bibinfo{editor}{Bottou, L.}, \bibinfo{editor}{Weinberger, K.Q.} (Eds.),
  \bibinfo{booktitle}{Advances in Neural Information Processing Systems}, pp.
  \bibinfo{pages}{1106--1114}.
\bibitem[{Laine and Aila(2017)}]{laine2016temporal}
\bibinfo{author}{Laine, S.}, \bibinfo{author}{Aila, T.}, \bibinfo{year}{2017}.
\newblock \bibinfo{title}{Temporal ensembling for semi-supervised learning},
  in: \bibinfo{booktitle}{5th International Conference on Learning
  Representations, {ICLR}}, \bibinfo{publisher}{OpenReview.net}.
\bibitem[{LeCun et~al.(2015)LeCun, Bengio and Hinton}]{lecun2015deep}
\bibinfo{author}{LeCun, Y.}, \bibinfo{author}{Bengio, Y.},
  \bibinfo{author}{Hinton, G.}, \bibinfo{year}{2015}.
\newblock \bibinfo{title}{Deep learning}.
\newblock \bibinfo{journal}{nature} \bibinfo{volume}{521},
  \bibinfo{pages}{436--444}.
\bibitem[{Lee(2013)}]{lee2013pseudo}
\bibinfo{author}{Lee, D.H.}, \bibinfo{year}{2013}.
\newblock \bibinfo{title}{Pseudo-label: The simple and efficient
  semi-supervised learning method for deep neural networks}, in:
  \bibinfo{booktitle}{Workshop on challenges in representation learning, ICML},
  p.~\bibinfo{pages}{2}.
\bibitem[{Li et~al.(2019a)Li, Chen, Ding, Zhu, Lu and Huang}]{li2019cycle}
\bibinfo{author}{Li, J.}, \bibinfo{author}{Chen, E.}, \bibinfo{author}{Ding,
  Z.}, \bibinfo{author}{Zhu, L.}, \bibinfo{author}{Lu, K.},
  \bibinfo{author}{Huang, Z.}, \bibinfo{year}{2019}a.
\newblock \bibinfo{title}{Cycle-consistent conditional adversarial transfer
  networks}, in: \bibinfo{booktitle}{Proceedings of the 27th ACM International
  Conference on Multimedia}, pp. \bibinfo{pages}{747--755}.
\bibitem[{Li et~al.(2020a)Li, Chen, Ding, Zhu, Lu and Shen}]{li2020maximum}
\bibinfo{author}{Li, J.}, \bibinfo{author}{Chen, E.}, \bibinfo{author}{Ding,
  Z.}, \bibinfo{author}{Zhu, L.}, \bibinfo{author}{Lu, K.},
  \bibinfo{author}{Shen, H.T.}, \bibinfo{year}{2020}a.
\newblock \bibinfo{title}{Maximum density divergence for domain adaptation}.
\newblock \bibinfo{journal}{IEEE transactions on pattern analysis and machine
  intelligence} .
\bibitem[{Li et~al.(2019b)Li, Jing, Lu, Zhu and Shen}]{li2019locality}
\bibinfo{author}{Li, J.}, \bibinfo{author}{Jing, M.}, \bibinfo{author}{Lu, K.},
  \bibinfo{author}{Zhu, L.}, \bibinfo{author}{Shen, H.T.},
  \bibinfo{year}{2019}b.
\newblock \bibinfo{title}{Locality preserving joint transfer for domain
  adaptation}.
\newblock \bibinfo{journal}{IEEE Transactions on Image Processing}
  \bibinfo{volume}{28}, \bibinfo{pages}{6103--6115}.
\bibitem[{Li and Zhang(2019)}]{Li2019SemiSupervisedDA}
\bibinfo{author}{Li, L.}, \bibinfo{author}{Zhang, Z.}, \bibinfo{year}{2019}.
\newblock \bibinfo{title}{Semi-supervised domain adaptation by covariance
  matching}.
\newblock \bibinfo{journal}{IEEE Transactions on Pattern Analysis and Machine
  Intelligence} \bibinfo{volume}{41}, \bibinfo{pages}{2724--2739}.
\bibitem[{Li et~al.(2019c)Li, Zhao, Sun, Cheng, Chen, Yan and
  Gao}]{Li2019WaveletKernelNetAI}
\bibinfo{author}{Li, T.}, \bibinfo{author}{Zhao, Z.}, \bibinfo{author}{Sun,
  C.}, \bibinfo{author}{Cheng, L.}, \bibinfo{author}{Chen, X.},
  \bibinfo{author}{Yan, R.}, \bibinfo{author}{Gao, R.}, \bibinfo{year}{2019}c.
\newblock \bibinfo{title}{Waveletkernelnet: An interpretable deep neural
  network for industrial intelligent diagnosis}.
\newblock \bibinfo{journal}{ArXiv} \bibinfo{volume}{abs/1911.07925}.
\bibitem[{Li et~al.(2020b)Li, Zhao, Sun, Yan and Chen}]{Li2020MultireceptiveFG}
\bibinfo{author}{Li, T.}, \bibinfo{author}{Zhao, Z.}, \bibinfo{author}{Sun,
  C.}, \bibinfo{author}{Yan, R.}, \bibinfo{author}{Chen, X.},
  \bibinfo{year}{2020}b.
\newblock \bibinfo{title}{Multi-receptive field graph convolutional networks
  for machine fault diagnosis}.
\newblock \bibinfo{journal}{IEEE Transactions on Industrial Electronics} ,
  \bibinfo{pages}{1--1}.
\bibitem[{Long et~al.(2015)Long, Cao, Wang and Jordan}]{long2015learning}
\bibinfo{author}{Long, M.}, \bibinfo{author}{Cao, Y.}, \bibinfo{author}{Wang,
  J.}, \bibinfo{author}{Jordan, M.I.}, \bibinfo{year}{2015}.
\newblock \bibinfo{title}{Learning transferable features with deep adaptation
  networks}, in: \bibinfo{editor}{Bach, F.R.}, \bibinfo{editor}{Blei, D.M.}
  (Eds.), \bibinfo{booktitle}{Proceedings of the 32nd International Conference
  on Machine Learning}, \bibinfo{publisher}{JMLR.org}. pp.
  \bibinfo{pages}{97--105}.
\bibitem[{Long et~al.(2018)Long, Cao, Wang and Jordan}]{long2018conditional}
\bibinfo{author}{Long, M.}, \bibinfo{author}{Cao, Z.}, \bibinfo{author}{Wang,
  J.}, \bibinfo{author}{Jordan, M.I.}, \bibinfo{year}{2018}.
\newblock \bibinfo{title}{Conditional adversarial domain adaptation}, in:
  \bibinfo{booktitle}{Advances in Neural Information Processing Systems}, pp.
  \bibinfo{pages}{1640--1650}.
\bibitem[{Long et~al.(2017)Long, Zhu, Wang and Jordan}]{long2017deep}
\bibinfo{author}{Long, M.}, \bibinfo{author}{Zhu, H.}, \bibinfo{author}{Wang,
  J.}, \bibinfo{author}{Jordan, M.I.}, \bibinfo{year}{2017}.
\newblock \bibinfo{title}{Deep transfer learning with joint adaptation
  networks}, in: \bibinfo{booktitle}{ICML}.
\bibitem[{Maaten and Hinton(2008)}]{maaten2008visualizing}
\bibinfo{author}{Maaten, L.v.d.}, \bibinfo{author}{Hinton, G.},
  \bibinfo{year}{2008}.
\newblock \bibinfo{title}{Visualizing data using t-sne}.
\newblock \bibinfo{journal}{Journal of machine learning research}
  \bibinfo{volume}{9}, \bibinfo{pages}{2579--2605}.
\bibitem[{Motiian et~al.(2017)Motiian, Piccirilli, Adjeroh and
  Doretto}]{motiian2017unified}
\bibinfo{author}{Motiian, S.}, \bibinfo{author}{Piccirilli, M.},
  \bibinfo{author}{Adjeroh, D.A.}, \bibinfo{author}{Doretto, G.},
  \bibinfo{year}{2017}.
\newblock \bibinfo{title}{Unified deep supervised domain adaptation and
  generalization}, in: \bibinfo{booktitle}{Proceedings of the IEEE
  International Conference on Computer Vision}, pp.
  \bibinfo{pages}{5715--5725}.
\bibitem[{Pan et~al.(2010)Pan, Tsang, Kwok and Yang}]{pan2010domain}
\bibinfo{author}{Pan, S.J.}, \bibinfo{author}{Tsang, I.W.},
  \bibinfo{author}{Kwok, J.T.}, \bibinfo{author}{Yang, Q.},
  \bibinfo{year}{2010}.
\newblock \bibinfo{title}{Domain adaptation via transfer component analysis}.
\newblock \bibinfo{journal}{IEEE Transactions on Neural Networks}
  \bibinfo{volume}{22}, \bibinfo{pages}{199--210}.
\bibitem[{Pan and Yang(2009)}]{pan2009survey}
\bibinfo{author}{Pan, S.J.}, \bibinfo{author}{Yang, Q.}, \bibinfo{year}{2009}.
\newblock \bibinfo{title}{A survey on transfer learning}.
\newblock \bibinfo{journal}{IEEE Transactions on knowledge and data
  engineering} \bibinfo{volume}{22}, \bibinfo{pages}{1345--1359}.
\bibitem[{Panareda~Busto and Gall(2017)}]{panareda2017open}
\bibinfo{author}{Panareda~Busto, P.}, \bibinfo{author}{Gall, J.},
  \bibinfo{year}{2017}.
\newblock \bibinfo{title}{Open set domain adaptation}, in:
  \bibinfo{booktitle}{Proceedings of the IEEE International Conference on
  Computer Vision}, pp. \bibinfo{pages}{754--763}.
\bibitem[{Purushotham et~al.(2017)Purushotham, Carvalho, Nilanon and
  Liu}]{SanjayVariational}
\bibinfo{author}{Purushotham, S.}, \bibinfo{author}{Carvalho, W.},
  \bibinfo{author}{Nilanon, T.}, \bibinfo{author}{Liu, Y.},
  \bibinfo{year}{2017}.
\newblock \bibinfo{title}{Variational recurrent adversarial deep domain
  adaptation}, in: \bibinfo{booktitle}{5th International Conference on Learning
  Representations}, \bibinfo{publisher}{OpenReview.net}.
\bibitem[{Qin et~al.(2020a)Qin, Wang, Ma, Yin, Wang and Fu}]{Qin2020OppositeSL}
\bibinfo{author}{Qin, C.}, \bibinfo{author}{Wang, L.}, \bibinfo{author}{Ma,
  Q.}, \bibinfo{author}{Yin, Y.}, \bibinfo{author}{Wang, H.},
  \bibinfo{author}{Fu, Y.}, \bibinfo{year}{2020}a.
\newblock \bibinfo{title}{Opposite structure learning for semi-supervised
  domain adaptation}.
\newblock \bibinfo{journal}{ArXiv} \bibinfo{volume}{abs/2002.02545}.
\bibitem[{Qin et~al.(2020b)Qin, Wang, Ma, Yin, Wang and Fu}]{qin2020opposite}
\bibinfo{author}{Qin, C.}, \bibinfo{author}{Wang, L.}, \bibinfo{author}{Ma,
  Q.}, \bibinfo{author}{Yin, Y.}, \bibinfo{author}{Wang, H.},
  \bibinfo{author}{Fu, Y.}, \bibinfo{year}{2020}b.
\newblock \bibinfo{title}{Opposite structure learning for semi-supervised
  domain adaptation}.
\newblock \bibinfo{journal}{arXiv preprint arXiv:2002.02545} .
\bibitem[{Rahman et~al.(2020)Rahman, Fookes, Baktash and
  Sridharan}]{Rahman2020OnMD}
\bibinfo{author}{Rahman, M.}, \bibinfo{author}{Fookes, C.},
  \bibinfo{author}{Baktash, M.}, \bibinfo{author}{Sridharan, S.},
  \bibinfo{year}{2020}.
\newblock \bibinfo{title}{On minimum discrepancy estimation for deep domain
  adaptation}.
\newblock \bibinfo{journal}{ArXiv} \bibinfo{volume}{abs/1901.00282}.
\bibitem[{Saenko et~al.(2010)Saenko, Kulis, Fritz and
  Darrell}]{saenko2010adapting}
\bibinfo{author}{Saenko, K.}, \bibinfo{author}{Kulis, B.},
  \bibinfo{author}{Fritz, M.}, \bibinfo{author}{Darrell, T.},
  \bibinfo{year}{2010}.
\newblock \bibinfo{title}{Adapting visual category models to new domains}, in:
  \bibinfo{booktitle}{ECCV}, \bibinfo{organization}{Springer}.
\bibitem[{Saito et~al.(2019)Saito, Kim, Sclaroff, Darrell and
  Saenko}]{saito2019semi}
\bibinfo{author}{Saito, K.}, \bibinfo{author}{Kim, D.},
  \bibinfo{author}{Sclaroff, S.}, \bibinfo{author}{Darrell, T.},
  \bibinfo{author}{Saenko, K.}, \bibinfo{year}{2019}.
\newblock \bibinfo{title}{Semi-supervised domain adaptation via minimax
  entropy}.
\newblock \bibinfo{journal}{CoRR} \bibinfo{volume}{abs/1904.06487}.
\bibitem[{Saito et~al.(2017)Saito, Ushiku and Harada}]{saito2017asymmetric}
\bibinfo{author}{Saito, K.}, \bibinfo{author}{Ushiku, Y.},
  \bibinfo{author}{Harada, T.}, \bibinfo{year}{2017}.
\newblock \bibinfo{title}{Asymmetric tri-training for unsupervised domain
  adaptation}, in: \bibinfo{booktitle}{Proceedings of the 34th International
  Conference on Machine Learning-Volume 70}, \bibinfo{organization}{JMLR. org}.
  pp. \bibinfo{pages}{2988--2997}.
\bibitem[{Saito et~al.(2018a)Saito, Ushiku, Harada and
  Saenko}]{saito2017adversarial}
\bibinfo{author}{Saito, K.}, \bibinfo{author}{Ushiku, Y.},
  \bibinfo{author}{Harada, T.}, \bibinfo{author}{Saenko, K.},
  \bibinfo{year}{2018}a.
\newblock \bibinfo{title}{Adversarial dropout regularization}, in:
  \bibinfo{booktitle}{6th International Conference on Learning
  Representationss}, \bibinfo{publisher}{OpenReview.net}.
\bibitem[{Saito et~al.(2018b)Saito, Watanabe, Ushiku and
  Harada}]{saito2018maximum}
\bibinfo{author}{Saito, K.}, \bibinfo{author}{Watanabe, K.},
  \bibinfo{author}{Ushiku, Y.}, \bibinfo{author}{Harada, T.},
  \bibinfo{year}{2018}b.
\newblock \bibinfo{title}{Maximum classifier discrepancy for unsupervised
  domain adaptation}, in: \bibinfo{booktitle}{Proceedings of the IEEE
  Conference on Computer Vision and Pattern Recognition}, pp.
  \bibinfo{pages}{3723--3732}.
\bibitem[{Shao et~al.(2018)Shao, Lan and Yuen}]{shao2018feature}
\bibinfo{author}{Shao, R.}, \bibinfo{author}{Lan, X.}, \bibinfo{author}{Yuen,
  P.C.}, \bibinfo{year}{2018}.
\newblock \bibinfo{title}{Feature constrained by pixel: Hierarchical
  adversarial deep domain adaptation}, in: \bibinfo{booktitle}{Proceedings of
  the 26th ACM international conference on Multimedia}, pp.
  \bibinfo{pages}{220--228}.
\bibitem[{Simonyan and Zisserman(2015)}]{simonyan2014very}
\bibinfo{author}{Simonyan, K.}, \bibinfo{author}{Zisserman, A.},
  \bibinfo{year}{2015}.
\newblock \bibinfo{title}{Very deep convolutional networks for large-scale
  image recognition}, in: \bibinfo{editor}{Bengio, Y.}, \bibinfo{editor}{LeCun,
  Y.} (Eds.), \bibinfo{booktitle}{3rd International Conference on Learning
  Representations}.
\bibitem[{Sun et~al.(2016)Sun, Feng and Saenko}]{sun2016return}
\bibinfo{author}{Sun, B.}, \bibinfo{author}{Feng, J.}, \bibinfo{author}{Saenko,
  K.}, \bibinfo{year}{2016}.
\newblock \bibinfo{title}{Return of frustratingly easy domain adaptation}, in:
  \bibinfo{booktitle}{Proceedings of the AAAI Conference on Artificial
  Intelligence}.
\bibitem[{Sun and Saenko(2016)}]{SunS16Deep}
\bibinfo{author}{Sun, B.}, \bibinfo{author}{Saenko, K.}, \bibinfo{year}{2016}.
\newblock \bibinfo{title}{Deep {CORAL:} correlation alignment for deep domain
  adaptation}, in: \bibinfo{editor}{Hua, G.}, \bibinfo{editor}{J{\'{e}}gou, H.}
  (Eds.), \bibinfo{booktitle}{Computer Vision - {ECCV} 2016 Workshops}, pp.
  \bibinfo{pages}{443--450}.
\bibitem[{Sun et~al.(2019)Sun, Wu, Luo, Gu, Yan and Du}]{Sun2019InformativeFS}
\bibinfo{author}{Sun, F.}, \bibinfo{author}{Wu, H.}, \bibinfo{author}{Luo, Z.},
  \bibinfo{author}{Gu, W.}, \bibinfo{author}{Yan, Y.}, \bibinfo{author}{Du,
  Q.}, \bibinfo{year}{2019}.
\newblock \bibinfo{title}{Informative feature selection for domain adaptation}.
\newblock \bibinfo{journal}{IEEE Access} \bibinfo{volume}{7},
  \bibinfo{pages}{142551--142563}.
\bibitem[{Szegedy et~al.(2014)Szegedy, Zaremba, Sutskever, Bruna, Erhan,
  Goodfellow and Fergus}]{DBLP:journals/corr/SzegedyZSBEGF13}
\bibinfo{author}{Szegedy, C.}, \bibinfo{author}{Zaremba, W.},
  \bibinfo{author}{Sutskever, I.}, \bibinfo{author}{Bruna, J.},
  \bibinfo{author}{Erhan, D.}, \bibinfo{author}{Goodfellow, I.J.},
  \bibinfo{author}{Fergus, R.}, \bibinfo{year}{2014}.
\newblock \bibinfo{title}{Intriguing properties of neural networks}, in:
  \bibinfo{editor}{Bengio, Y.}, \bibinfo{editor}{LeCun, Y.} (Eds.),
  \bibinfo{booktitle}{2nd International Conference on Learning
  Representations}.
\bibitem[{Tang and Jia(2019)}]{tang2019discriminative}
\bibinfo{author}{Tang, H.}, \bibinfo{author}{Jia, K.}, \bibinfo{year}{2019}.
\newblock \bibinfo{title}{Discriminative adversarial domain adaptation}.
\newblock \bibinfo{journal}{arXiv preprint arXiv:1911.12036} .
\bibitem[{Tarvainen and Valpola(2017)}]{DBLP:conf/iclr/TarvainenV17}
\bibinfo{author}{Tarvainen, A.}, \bibinfo{author}{Valpola, H.},
  \bibinfo{year}{2017}.
\newblock \bibinfo{title}{Mean teachers are better role models: Weight-averaged
  consistency targets improve semi-supervised deep learning results}, in:
  \bibinfo{booktitle}{5th International Conference on Learning
  Representations}, \bibinfo{publisher}{OpenReview.net}.
\bibitem[{Tzeng et~al.(2014)Tzeng, Hoffman, Zhang, Saenko and
  Darrell}]{tzeng2014deep}
\bibinfo{author}{Tzeng, E.}, \bibinfo{author}{Hoffman, J.},
  \bibinfo{author}{Zhang, N.}, \bibinfo{author}{Saenko, K.},
  \bibinfo{author}{Darrell, T.}, \bibinfo{year}{2014}.
\newblock \bibinfo{title}{Deep domain confusion: Maximizing for domain
  invariance}.
\newblock \bibinfo{journal}{arXiv preprint arXiv:1412.3474} .
\bibitem[{Venkateswara et~al.(2017)Venkateswara, Eusebio, Chakraborty and
  Panchanathan}]{venkateswara2017deep}
\bibinfo{author}{Venkateswara, H.}, \bibinfo{author}{Eusebio, J.},
  \bibinfo{author}{Chakraborty, S.}, \bibinfo{author}{Panchanathan, S.},
  \bibinfo{year}{2017}.
\newblock \bibinfo{title}{Deep hashing network for unsupervised domain
  adaptation}, in: \bibinfo{booktitle}{Proceedings of the IEEE Conference on
  Computer Vision and Pattern Recognition}, pp. \bibinfo{pages}{5018--5027}.
\bibitem[{Wu et~al.(2020a)Wu, Yan, Lin, Yang, Ng and Wu}]{Wu2020IterativeRF}
\bibinfo{author}{Wu, H.}, \bibinfo{author}{Yan, Y.}, \bibinfo{author}{Lin, G.},
  \bibinfo{author}{Yang, M.}, \bibinfo{author}{Ng, M.}, \bibinfo{author}{Wu,
  Q.}, \bibinfo{year}{2020}a.
\newblock \bibinfo{title}{Iterative refinement for multi-source visual domain
  adaptation}.
\newblock \bibinfo{journal}{IEEE Transactions on Knowledge and Data
  Engineering} , \bibinfo{pages}{1--1}.
\bibitem[{Wu et~al.(2020b)Wu, Yan, Ye, Ng and Wu}]{Wu2020GeometricKE}
\bibinfo{author}{Wu, H.}, \bibinfo{author}{Yan, Y.}, \bibinfo{author}{Ye, Y.},
  \bibinfo{author}{Ng, M.}, \bibinfo{author}{Wu, Q.}, \bibinfo{year}{2020}b.
\newblock \bibinfo{title}{Geometric knowledge embedding for unsupervised domain
  adaptation}.
\newblock \bibinfo{journal}{Knowl. Based Syst.} \bibinfo{volume}{191},
  \bibinfo{pages}{105155}.
\bibitem[{Xie et~al.(2018)Xie, Zheng, Chen and Chen}]{xie2018learning}
\bibinfo{author}{Xie, S.}, \bibinfo{author}{Zheng, Z.}, \bibinfo{author}{Chen,
  L.}, \bibinfo{author}{Chen, C.}, \bibinfo{year}{2018}.
\newblock \bibinfo{title}{Learning semantic representations for unsupervised
  domain adaptation}, in: \bibinfo{editor}{Dy, J.G.}, \bibinfo{editor}{Krause,
  A.} (Eds.), \bibinfo{booktitle}{Proceedings of the 35th International
  Conference on Machine Learning}, \bibinfo{publisher}{{PMLR}}. pp.
  \bibinfo{pages}{5419--5428}.
\bibitem[{Xu et~al.(2019)Xu, Li, Yang and Lin}]{xu2019larger}
\bibinfo{author}{Xu, R.}, \bibinfo{author}{Li, G.}, \bibinfo{author}{Yang, J.},
  \bibinfo{author}{Lin, L.}, \bibinfo{year}{2019}.
\newblock \bibinfo{title}{Larger norm more transferable: An adaptive feature
  norm approach for unsupervised domain adaptation}, in:
  \bibinfo{booktitle}{Proceedings of the IEEE International Conference on
  Computer Vision}, pp. \bibinfo{pages}{1426--1435}.
\bibitem[{Yan et~al.(2018)Yan, Li, Wu, Min, Tan and
  Wu}]{Yan2018SemiSupervisedOT}
\bibinfo{author}{Yan, Y.}, \bibinfo{author}{Li, W.}, \bibinfo{author}{Wu, H.},
  \bibinfo{author}{Min, H.}, \bibinfo{author}{Tan, M.}, \bibinfo{author}{Wu,
  Q.}, \bibinfo{year}{2018}.
\newblock \bibinfo{title}{Semi-supervised optimal transport for heterogeneous
  domain adaptation}, in: \bibinfo{booktitle}{IJCAI}.
\bibitem[{Yang et~al.(2020a)Yang, Zou, Zhou, Zeng and Xie}]{yang2020mind}
\bibinfo{author}{Yang, J.}, \bibinfo{author}{Zou, H.}, \bibinfo{author}{Zhou,
  Y.}, \bibinfo{author}{Zeng, Z.}, \bibinfo{author}{Xie, L.},
  \bibinfo{year}{2020}a.
\newblock \bibinfo{title}{Mind the discriminability: Asymmetric adversarial
  domain adaptation}, in: \bibinfo{booktitle}{European Conference on Computer
  Vision}, \bibinfo{organization}{Springer}. pp. \bibinfo{pages}{589--606}.
\bibitem[{Yang et~al.(2020b)Yang, Wang, Gao, Shrivastava, Weinberger, Chao and
  Lim}]{Yang2020DeepCW}
\bibinfo{author}{Yang, L.}, \bibinfo{author}{Wang, Y.}, \bibinfo{author}{Gao,
  M.}, \bibinfo{author}{Shrivastava, A.}, \bibinfo{author}{Weinberger, K.Q.},
  \bibinfo{author}{Chao, W.L.}, \bibinfo{author}{Lim, S.N.},
  \bibinfo{year}{2020}b.
\newblock \bibinfo{title}{Deep co-training with task decomposition for
  semi-supervised domain adaptation.}
\newblock \bibinfo{journal}{arXiv: Computer Vision and Pattern Recognition} .
\bibitem[{Zhang et~al.(a)Zhang, Bengio, Hardt, Recht and
  Vinyals}]{zhang2016understanding}
\bibinfo{author}{Zhang, C.}, \bibinfo{author}{Bengio, S.},
  \bibinfo{author}{Hardt, M.}, \bibinfo{author}{Recht, B.},
  \bibinfo{author}{Vinyals, O.}, a.
\newblock \bibinfo{title}{Understanding deep learning requires rethinking
  generalization}, in: \bibinfo{booktitle}{5th International Conference on
  Learning Representations}.
\bibitem[{Zhang et~al.(b)Zhang, Liu, Long and Jordan}]{zhang2019bridging}
\bibinfo{author}{Zhang, Y.}, \bibinfo{author}{Liu, T.}, \bibinfo{author}{Long,
  M.}, \bibinfo{author}{Jordan, M.I.}, b.
\newblock \bibinfo{title}{Bridging theory and algorithm for domain adaptation},
  in: \bibinfo{editor}{Chaudhuri, K.}, \bibinfo{editor}{Salakhutdinov, R.}
  (Eds.), \bibinfo{booktitle}{Proceedings of the 36th International Conference
  on Machine Learning}.
\bibitem[{Zhang et~al.(2019)Zhang, Tang, Jia and Tan}]{zhang2019domain}
\bibinfo{author}{Zhang, Y.}, \bibinfo{author}{Tang, H.}, \bibinfo{author}{Jia,
  K.}, \bibinfo{author}{Tan, M.}, \bibinfo{year}{2019}.
\newblock \bibinfo{title}{Domain-symmetric networks for adversarial domain
  adaptation}, in: \bibinfo{booktitle}{Proceedings of the IEEE Conference on
  Computer Vision and Pattern Recognition}, pp. \bibinfo{pages}{5031--5040}.
\bibitem[{Zhang et~al.(2020)Zhang, Wei, Wu, Zhao, Niu, Huang and
  Tan}]{Zhang2020CollaborativeUD}
\bibinfo{author}{Zhang, Y.}, \bibinfo{author}{Wei, Y.}, \bibinfo{author}{Wu,
  Q.}, \bibinfo{author}{Zhao, P.}, \bibinfo{author}{Niu, S.},
  \bibinfo{author}{Huang, J.}, \bibinfo{author}{Tan, M.}, \bibinfo{year}{2020}.
\newblock \bibinfo{title}{Collaborative unsupervised domain adaptation for
  medical image diagnosis}.
\newblock \bibinfo{journal}{IEEE Transactions on Image Processing}
  \bibinfo{volume}{29}, \bibinfo{pages}{7834--7844}.
\bibitem[{Zou et~al.(2019)Zou, Yu, Liu, Kumar and Wang}]{zou2019confidence}
\bibinfo{author}{Zou, Y.}, \bibinfo{author}{Yu, Z.}, \bibinfo{author}{Liu, X.},
  \bibinfo{author}{Kumar, B.}, \bibinfo{author}{Wang, J.},
  \bibinfo{year}{2019}.
\newblock \bibinfo{title}{Confidence regularized self-training}, in:
  \bibinfo{booktitle}{Proceedings of the IEEE International Conference on
  Computer Vision}, pp. \bibinfo{pages}{5982--5991}.
\bibitem[{Zou et~al.(2018)Zou, Yu, Vijaya~Kumar and Wang}]{zou2018domain}
\bibinfo{author}{Zou, Y.}, \bibinfo{author}{Yu, Z.},
  \bibinfo{author}{Vijaya~Kumar, B.}, \bibinfo{author}{Wang, J.},
  \bibinfo{year}{2018}.
\newblock \bibinfo{title}{Unsupervised domain adaptation for semantic
  segmentation via class-balanced self-training}, in:
  \bibinfo{booktitle}{Proceedings of the European conference on computer vision
  (ECCV)}, pp. \bibinfo{pages}{289--305}.

\end{thebibliography}

\end{document}